\documentclass[sigconf, preprint, nonacm]{acmart}
\AtBeginDocument{
  }

\usepackage{multirow}
\usepackage[table]{xcolor}
\usepackage{graphicx} 
\usepackage{subcaption}
\usepackage{caption} 
\usepackage{adjustbox}
\usepackage{placeins}

\usepackage{xcolor}
\usepackage{listings}

\lstdefinestyle{promptbox}{
  basicstyle=\ttfamily\footnotesize,
  breaklines=true,
  columns=fullflexible,
  backgroundcolor=\color{gray!10},
  frame=single,
  rulecolor=\color{gray!40},
  xleftmargin=0.4em,
  xrightmargin=0.4em,
  aboveskip=0.4em,
  belowskip=0.2em,
  mathescape=true,
  showstringspaces=false
}

\setcopyright{none}
\copyrightyear{}
\acmYear{}
\acmDOI{}
\acmISBN{}

\begin{document}

\title{GNNVerifier: Graph-based Verifier for LLM Task Planning}

\author{Yu Hao}
\affiliation{
  \institution{Beijing University of Posts and Telecommunications}
  \city{Beijing}
  \country{China}}
\email{haoyuu@bupt.edu.cn}

\author{Qiuyu Wang}
\affiliation{
  \institution{Beijing University of Posts and Telecommunications}
  \city{Beijing}
  \country{China}}
\email{autumn@bupt.edu.cn}

\author{Cheng Yang}
\authornote{Corresponding author.}
\affiliation{
  \institution{Beijing University of Posts and Telecommunications}
  \city{Beijing}
  \country{China}}
\email{yangcheng@bupt.edu.cn}

\author{Yawen Li}
\affiliation{
  \institution{Beijing University of Posts and Telecommunications}
  \city{Beijing}
  \country{China}}
\email{warmly0716@126.com}

\author{Zhiqiang Zhang}
\affiliation{
  \institution{Ant Group}
  \city{Beijing}
  \country{China}}
\email{lingyao.zzq@antfin.com}

\author{Chuan Shi}
\affiliation{
  \institution{Beijing University of Posts and Telecommunications}
  \city{Beijing}
  \country{China}}
\email{shichuan@bupt.edu.cn}

\renewcommand{\shortauthors}{Trovato et al.}

\begin{abstract}
Large language models (LLMs) facilitate the development of autonomous agents. As a core component of such agents, task planning aims to decompose complex natural language requests into concrete, solvable sub-tasks. Since LLM-generated plans are frequently prone to hallucinations and sensitive to long-context prom-pts, recent research has introduced plan verifiers to identify and correct potential flaws. However, most existing approaches still rely on an LLM as the verifier via additional prompting for plan review or self-reflection. LLM-based verifiers can be misled by plausible narration and struggle to detect failures caused by structural relations across steps, such as type mismatches, missing intermediates, or broken dependencies. To address these limitations, we propose a graph-based verifier for LLM task planning. Specifically, the proposed method has four major components: Firstly, we represent a plan as a directed graph with enriched attributes, where nodes denote sub-tasks and edges encode execution order and dependency constraints. Secondly, a graph neural network (GNN) then performs structural evaluation and diagnosis, producing a graph-level plausibility score for plan acceptance as well as node/edge-level risk scores to localize erroneous regions. Thirdly, we construct controllable perturbations from ground truth plan graphs, and automatically generate training data with fine-grained annotations. Finally, guided by the feedback from our GNN verifier, we enable an LLM to conduct local edits (e.g., tool replacement or insertion) to correct the plan when the graph-level score is insufficient. Extensive experiments across diverse datasets, backbone LLMs, and planners demonstrate that our GNNVerifier achieves significant gains in improving plan quality. Our data and code is available at https://github.com/BUPT-GAMMA/GNNVerifier.
\end{abstract}




\received{20 February 2007}
\received[revised]{12 March 2009}
\received[accepted]{5 June 2009}

\maketitle

\section{Introduction}

Large language models (LLMs) have achieved remarkable progress in natural language understanding and complex reasoning, gradually evolving from text generators into general purpose agents~\cite{brown2020language,wei2022chain}. In this background, task planning has become a core component of such agents: it aims to decompose a complex user request expressed in natural language into a sequence of concrete, solvable sub-tasks and to organize them in an appropriate order~\cite{wu2023autogen,shen2023hugginggpt,ahn2022can,wu2024toolplanner}. Task planning has been widely adopted in practical tool-augmented agent scenarios, such as retrieval augmented question answering~\cite{lewis2020retrieval}, office workflow automation~\cite{wang2024officebench}, and multimodal content generation and editing~\cite{wu2023visual,yang2023mm}.

Existing task planning methods often rely on LLMs to generate multi-step plans via prompting~\cite{wang2023plan,shen2023hugginggpt,wu2024toolplanner}. A common practice is to pack tool documentation, constraint rules, and exemplars into the context, so that the model performs task decomposition through in-context learning and outputs a step sequence and tool-calling chain~\cite{wu2024toolplanner,shen2023hugginggpt}. But these methods typically lead to increasingly long contexts, which can dilute attention and increase hallucinations, causing plans that look plausible but are non-executable or internally inconsistent~\cite{ji2023survey,kambhampati2024llms}.

\begin{figure}[t]
  \centering
  \includegraphics[width=\linewidth]{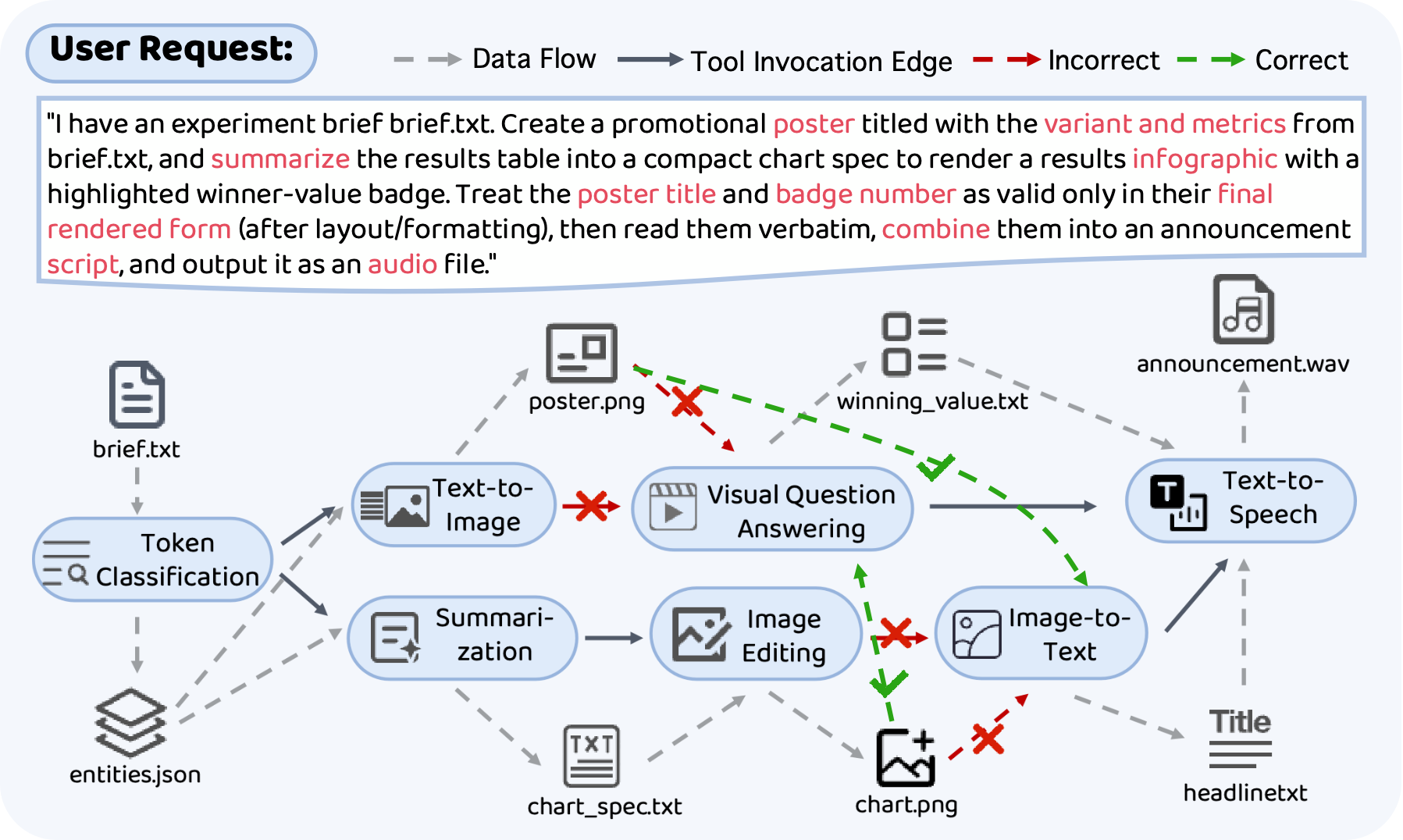}
  \caption{An illustrative example of structural inconsistencies in LLM-generated plans. Although all steps are fluent and seemingly plausible, two image artifacts (poster.png and chart.png) are mistakenly input into the opposite downstream tools from their originally intended ones, which is difficult to detect by reading steps in isolation.}
  \label{fig:intro}
\end{figure}

In this background, prior work has introduced plan verifiers to improve planning reliability~\cite{lee2025veriplan,ho2025verilogcoder,hariharan2025plan}. A plan verifier independently evaluates a generated plan for a given user request, determines whether it should be accepted, and localizes potential issues to trigger correction. Most existing approaches still rely on an LLM as the verifier via additional prompting for plan review or self-reflection~\cite{madaan2023self,shinn2023reflexion}. However, such methods have clear limitations. As shown in Figure~\ref{fig:intro}, on the one hand, LLM-based verifiers can be misled by fluent step descriptions, mistaking plausible narration for correct execution~\cite{valmeekam2023planning,gu2024survey}. On the other hand, many failures arise from structural relations across steps (e.g., type mismatches, missing critical intermediate steps, or broken dependencies), which are difficult to detect and pinpoint by reading steps in isolation~\cite{valmeekam2023planning,wu2024can}.

Based on these insights, we propose a graph-based verifier for LLM task planning. We first represent a plan as a directed plan graph~\cite{liu2024toolnet,lumer2025graph} with enriched attributes, where nodes correspond to sub-tasks and edges encode execution order and dependency constraints. Then we use a graph neural network (GNN) to conduct structural evaluation and diagnosis. Our model outputs (i) a graph-level plausibility score to decide whether to accept the current plan, and (ii) node- and edge-level risk scores to localize erroneous regions. Moreover, since existing data lack fine-grained annotations for erroneous plan graphs, we construct controllable perturbations from ground truth plan graphs and automatically generate supervision signals. This enables the verifier to learn to distinguish correct plans from incorrect ones while marking high-risk regions on the graph. Finally, we revise the plan graph based on feedback from the GNN verifier: when the graph-level score is not sufficiently high, an LLM is allowed to perform local edits at high-risk locations based on tool replacement or insertion. Overall, our GNN verifier can leverage the structural information in the plan graph to uncover potential issues and help the LLM correct them. Experimental results show that, compared with the best baseline, our GNNVerifier has 2.13\%/9.22\%/15.96\% relative improvements of node/edge/graph-level metrics, respectively. We summarize our main contributions as follows:
\begin{itemize}
\item We innovatively propose to use graph-based verifiers instead of LLM-based ones to effectively identify structural issues in LLM-generated task plans.

\item By modeling LLM-generated plans as directed, attributed graphs, we propose a GNN verifier to predict graph-level plausibility scores and node/edge-level risk scores. The scores are then utilized for plan correction.

\item We systematically compare against state-of-the-art baselines across diverse combinations of datasets, backbone LLMs, and planners. Results show that our GNNVerifier achieves significant gains in improving plan quality.
\end{itemize}

\section{Related Work}
\subsection{Planning with Large Language Models}
Existing paradigms in LLM-based planning can generally be categorized into two primary directions: Intrinsic LLM Planning and External Planner Integration.

\textbf{Intrinsic LLM Planning} refers to approaches where the LLM utilizes its own knowledge and capabilities to generate sub-task sequences. Typical methods like CoT~\cite{wei2022chain}, ReAct~\cite{yao2022react}, and HuggingGPT~\cite{shen2023hugginggpt} break down complex tasks into manageable subgoals and plan for each successively. Following this line, Adapt~\cite{prasad2024adapt} further dynamically adjusts the planning process based on both task complexity and the LLM's inherent capabilities. Alternatively,  ToT~\cite{yao2023tree}, GoT~\cite{besta2024graph} and CoT-SC~\cite{wang2022self} explore the solution space by generating multiple candidate trajectories. In these frameworks, LLMs also function as evaluators to assess these trajectories and identify the most effective plan.

\textbf{External Planner Integration} adopts a symbolic component or a small neural network to handle intricate constraints, assisting the LLM for better planning performance. For example, some works~\cite{liu2023llm+, guan2023leveraging, dagan2023dynamic} use LLMs to translate natural language problems into Planning Domain Definition Language (PDDL), which are then solved by classical symbolic solvers. Besides, sub-tasks in planning can form a graph where nodes represent tasks and edges represent dependencies. Empirical investigation of GNN4Plan~\cite{wu2024can} indicates that planning failures can be ascribed to the LLMs' inefficacy in accurately discerning the structure of the plan graph. Consequently, GNNs can assist LLMs by effectively handling these constraints.

\subsection{Verifiers for Large Language Models}
Verifiers can deliver meaningful feedback to LLMs. Early research on verifiers predominantly concentrated on mathematical tasks and code generation, where tasks are objectively verifiable through unit tests or final answers. In these methods, verifiers can be categorized into Outcome Reward Models (ORMs)~\cite{cobbe2021training}, which assess the correctness of the final answer, and Process Reward Models (PRMs)~\cite{lightman2023let}, which evaluate the reasoning trajectory step-by-step.

Recent studies have extended the application of verifiers to more challenging domains, such as open-domain question answering and plan generation. For instance, VersaPRM~\cite{anonymous2026unified} introduces a general PRM trained on synthetic Chain-of-Thought (CoT) traces and counterfactual variants. Since these areas are often characterized by high subjectivity and inherent difficulties in verification~\cite{xiong2025rag}, generative verifiers have also been employed in these fields to generate natural language critiques that assist in self-correction~\cite{madaan2023self}.

In the domain of planning, a few verifiers have emerged to provide modification feedback to initial plans~\cite{kambhampati2024position}. For example, VeriPlan~\cite{lee2025veriplan} employs model checking against LLM outputs, incorporating multiple user-control features. Another work iteratively validates sub-tasks against module descriptions, refining the plan until full consistency is achieved~\cite{ho2025verilogcoder}.

However, existing methods often overlook the structural dependencies within the plan graph, which are pivotal for robust planning performance~\cite{wu2024can}. To address this limitation, we propose a graph-based verifier that evaluates the initial plan by generating node-, edge-, and graph-level scores, thereby providing structural-aware feedback for plan correction.

\begin{figure*}[t]
  \centering
  \setlength{\tabcolsep}{6pt}
  \begin{tabular}{@{}c@{}}
    \includegraphics[width=1.0\linewidth]{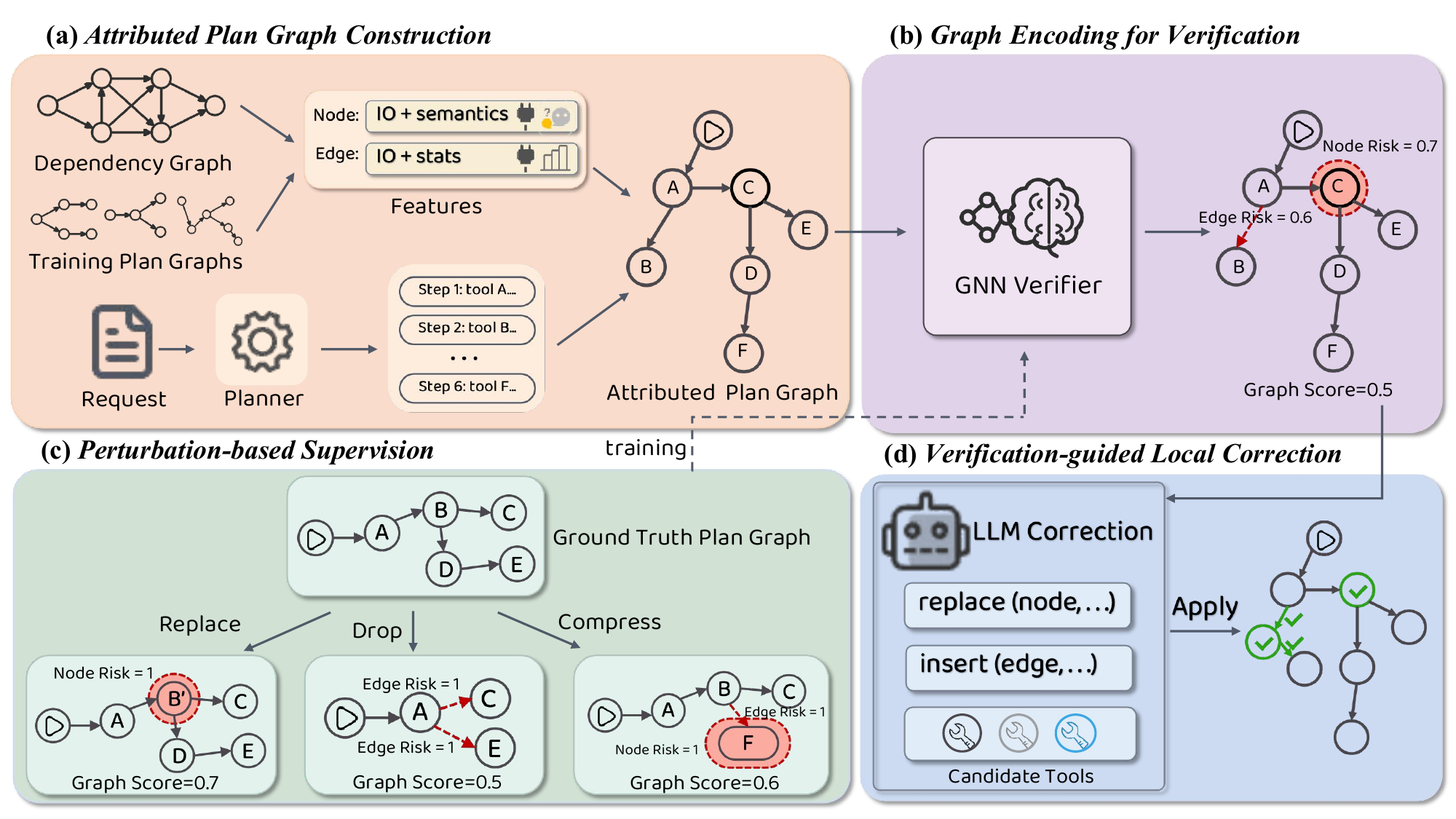}  
    \\[0.4em]
  \end{tabular}
  \caption{The overall framework of our proposed GNNVerifier with four main components.} 
  \label{fig:method} 
\end{figure*}

\section{Methodology}
\label{sec:method}

\subsection{Problem Formalization}
\label{sec:formalization}
In this subsection, we first introduce the LLM-based task planing, and then formalize the problem of graph-based plan verification.

\textbf{Task Planning.} Given a natural language user request $r$ and the tool set $\mathcal{T}$, task planning aims to decompose $r$ into a sequence of concrete, solvable sub-tasks and arrange them in an appropriate order.
Following common settings in tool-augmented agents~\cite{schick2023toolformer,shen2023hugginggpt,shen2024taskbench}, we represent a plan as (i) a step text sequence $S_r=\langle s_1,\ldots,s_m\rangle$ and (ii) an aligned tool trajectory $\tau_r=\langle t_1,\ldots,t_m\rangle$ with $t_i\in\mathcal{T}$.
Here, $s_i$ describes the intent of the $i$-th sub-task, and $t_i$ is the tool selected to execute that step. 
Many LLM-based planners~\cite{shen2023hugginggpt,shen2024taskbench,wu2024can,wang2023plan} generate such sequences by first decomposing the request and then selecting appropriate tools. We do not assume any specific internal mechanism of the planner, and only require that its output can be converted into the above form.

\textbf{Dependency Graph.} Following previous work~\cite{shen2024taskbench,wu2024can}, we use a dependency graph to represent the tool set and their relations. Formally, the dependency graph is defined as $G_{\text{tool}}=(\mathcal{T},\mathcal{D})$, where $\mathcal{T}=\{t_1,t_2,\ldots,t_{|\mathcal{T}|}\}$ denotes the set of available tools.
Each tool $t\in\mathcal{T}$ is associated with a textual description $\mathrm{desc}(t)$ and input/output type sets $\mathrm{in}(t)$ and $\mathrm{out}(t)$.
$\mathcal{D}\subseteq \mathcal{T}\times \mathcal{T}$ is a set of directed edges; for any $(t_u,t_v)\in\mathcal{D}$, the interface type compatibility constraint holds:
$\mathrm{out}(t_u)\cap \mathrm{in}(t_v)\neq\emptyset.$
Equivalently, $\mathcal{D}$ induces directed tool neighborhoods
$\mathcal{N}^{\text{tool}}_{\mathrm{out}}(t)=\{t'\in\mathcal{T}\mid (t,t')\in\mathcal{D}\}$
and
$\mathcal{N}^{\text{tool}}_{\mathrm{in}}(t)=\{t'\in\mathcal{T}\mid (t',t)\in\mathcal{D}\}$.
This graph captures the minimal executability requirement of tool interfaces and serves as the basis for subsequent plan representations and constraint construction.

\textbf{Plan Graph.} Given the planning result $(S_r,\tau_r)$ for request $r$, we convert it into a directed plan graph $G_r=(V_r,E_r)$.
Each node $v_i\in V_r$ corresponds to a tool invocation instance and is denoted as $v_i=(t_i,s_i)$, where $t_i\in\mathcal{T}$ is the invoked tool and $s_i$ is the natural language description of the step. A directed edge $(u,v)\in E_r$ encodes the execution order and dependency constraints between tool invocations. To uniformly handle missing prerequisite steps, we introduce a virtual start node \textsc{Start} and connect it to all tool nodes with zero in-degree.

\textbf{Graph-based Plan Verification.} A graph-based plan verifier assesses the feasibility and plausibility of a plan produced by a planner under a given user request, and outputs feedback signals that can be used to improve the plan.
Formally, given request $r$ and its plan graph $G_r$, a verifier can be abstracted as a mapping
\begin{equation}
\label{eq:verifier-map}
\mathcal{V}:\ (r,\,G_r)\ \mapsto\ o_r,
\end{equation}
where $o_r$ denotes the verification signals.
These signals may be a global quality assessment, optionally augmented with localized diagnostic feedback (e.g., indicating potentially problematic steps / dependencies), or natural language critiques and suggested edits.
The verifier output is typically used as conditional input to a subsequent corrector (e.g., an LLM-based corrector) to produce an improved plan $G'_r$.

\subsection{Framework Overview}
\label{sec:overview}
As shown in Figure~\ref{fig:method}, our method learns a graph-based verifier with four components:
(1) \emph{Attributed plan graph construction} converts the planner output into a directed plan graph with enriched node/edge features;
(2) \emph{Graph encoding for verification} applies a GNN to predict a graph-level plausibility score and node/edge-level risks;
(3) \emph{Perturbation-based supervision} perturbs ground truth graphs to derive graph/node/edge-level training signals;
(4) \emph{Verification-guided local correction} constrains an LLM to perform replacements/insertions on high-risk regions for correction.

\subsection{Attributed Plan Graph Construction}
\label{sec:plan-graph-const}
In this subsection, we design rich node and edge features to construct attributed plan graphs for better verification. 
\subsubsection{Preprocessing}
To enrich the features of plan graph $G_r=(V_r,E_r)$, we precompute two terms from the dependency graph $G_{\text{tool}}=(\mathcal{T},\mathcal{D})$ and the training data: (i) for each tool $t$, we encode its description $\mathrm{desc}(t)$ into a semantic embedding $\mathbf{e}(t)=\mathrm{Enc}(\mathrm{desc}(t))$, and compute the Top-$K$ neighborhood $\mathcal{N}_K(t)\subset\mathcal{T}$ ranked by cosine similarity; (ii) for any length-$n$ path $(t_1\!\rightarrow t_2\!\rightarrow\cdots\!\rightarrow t_n)$, we denote its occurrence count in the training set as $f_n(t_1,\ldots,t_n)$ with $n\in\{2,3,4\}$.

\subsubsection{Node Features}
Each node is a tool invocation instance $v_i=(t_i,s_i)$, where $t_i\in\mathcal{T}$ is the selected tool and $s_i$ is the step text.
We construct node features by integrating: (i) tool semantics $\mathbf{e}(t_i)=\mathrm{Enc}(\mathrm{desc}(t_i))$ and step semantics $\mathbf{e}(s_i)=\mathrm{Enc}(s_i)$; (ii) multi-hot encodings of tool I/O types $\mathbf{x}_{\text{in}}(t_i)$ and $\mathbf{x}_{\text{out}}(t_i)$; and (iii) a lightweight step--tool alignment scorer $g(s,t)$ to measure intent--tool consistency.
Concretely, for each ground truth node $v_i=(t_i,s_i)$ in training data, we build a candidate set $\{t_i\}\cup \mathcal{N}_K(t_i)$, treat $t_i$ as the positive and the remaining candidates as negatives, and compute a logit via
\begin{equation}
g(s_i,t)=\mathrm{MLP}_{\text{align}}\big([\mathbf{e}(s_i);\mathbf{e}(t)]\big).
\label{eq:align-scorer}
\end{equation}
We train $g$ with a softmax cross-entropy loss over $\{t_i\}\cup \mathcal{N}_K(t_i)$, encouraging fine-grained discrimination among similar tools.
We then define the step--tool alignment feature
\begin{equation}
\Delta_i = g(s_i,t_i)- \max_{t'\in \mathcal{N}_K(t_i)} g(s_i,t').
\label{eq:delta-align}
\end{equation}
A smaller $\Delta_i$ indicates that the step text is harder to distinguish among similar tools.
We concatenate these features and obtain the node embedding via an MLP (multi-layer perceptron), i.e., $x_{v_i}=\mathrm{MLP}_{\text{node}}(\cdot)$.
The \textsc{Start} node is represented with a learnable embedding.

\subsubsection{Edge Features}
For each directed edge $(u,v)\in E_r$, we construct three features.

\textbf{I/O compatibility.} Based on type constraints from $G_{\text{tool}}$, we define
  \begin{equation}
  \mathrm{compat}(u,v)=\frac{|\mathrm{out}(t_u)\cap \mathrm{in}(t_v)|}{\max(|\mathrm{in}(t_v)|,1)},
  \label{eq:compat}
  \end{equation}
  which measures whether two adjacent tools are connectable at the interface level and how strong the connection is.

\textbf{Pairwise co-occurrence.}
  Using length-2 statistics, we set the transition strength as $\log(1+f_2(t_u,t_v))$, indicating whether this adjacency is common.

  \textbf{Multi-step relationship.}
  To reflect whether transitioning from $u$ to $v$ typically requires intermediate tools, we define
  \begin{equation}  m(u,v)=\max_{\pi:\,t_u\rightarrow\cdots\rightarrow t_v,\ |\pi|=n}\ \log(1+f_n(\pi)),\quad n\in\{3,4\},
  \label{eq:motif}
  \end{equation}
  where $\pi$ is a length-$n$ directed tool sequence from $t_u$ to $t_v$.
  A larger $m(u,v)$ suggests that $t_u\!\rightarrow t_v$ may correspond to an unreliable shortcut (e.g., over-compression or missing steps).

We concatenate these features and obtain the edge embedding via an MLP, i.e., $x_{uv}=\mathrm{MLP}_{\text{edge}}(\cdot)$.
For the virtual edges from $\textsc{Start}$, we use a learnable edge embedding.

\subsection{Graph Encoding for Verification}
\label{sec:encoding}
In this subsection, we present how to encode the attributed plan graph for graph/node/edge-level scoring.
\subsubsection{GNN-based Representation}
\label{sec:graph-encoding}
We encode the plan graph with a directed, edge-aware GNN conditioned on the request embedding
$\mathbf{e}(r)=\mathrm{Enc}(r)$. 
Let $\mathbf{h}_v^{(0)}=\mathbf{x}_v$, at layer $\ell$, we aggregate messages from incoming and outgoing neighbors separately:
\begin{equation}
\begin{aligned}
\mathbf{m}_{v,\mathrm{in}}^{(\ell)}
&=
\sum_{u \in \mathcal{N}_{\mathrm{in}}(v)}
\phi_{\mathrm{in}}^{(\ell)}\Big([\mathbf{h}_u^{(\ell)};\ \mathbf{x}_{uv};\ \mathbf{e}(r)]\Big),\\
\mathbf{m}_{v,\mathrm{out}}^{(\ell)}
&=
\sum_{w \in \mathcal{N}_{\mathrm{out}}(v)}
\phi_{\mathrm{out}}^{(\ell)}\Big([\mathbf{h}_w^{(\ell)};\ \mathbf{x}_{vw};\ \mathbf{e}(r)]\Big),
\end{aligned}
\label{eq:msg-agg}
\end{equation}
and update node representations by
\begin{equation}
\label{eq:msg-update}
\mathbf{h}_v^{(\ell+1)}=
\mathrm{MLP}^{(\ell)}\!\Big((1+\epsilon^{(\ell)})\mathbf{h}_v^{(\ell)}+
\mathbf{m}_{v,\mathrm{in}}^{(\ell)}+\mathbf{m}_{v,\mathrm{out}}^{(\ell)}\Big),
\end{equation}
where $\phi^{(\ell)}$ is a small MLP and $\epsilon^{(\ell)}$ is learnable.
After $L$ layers, we obtain final node representations $\{\mathbf{h}_v\}_{v\in V_r}$ and a graph representation
$\mathbf{h}_G=\mathrm{READOUT}(\{\mathbf{h}_v\}_{v\in V_r})$.

\subsubsection{Verifier Outputs}
\label{sec:verifier-outputs}
Given request $r$ and plan graph $G_r=(V_r,E_r)$, a graph-based plan verifier is expected to provide both a global quality assessment and localized risk diagnosis to support subsequent local correction. We let the verifier output three probabilistic signals: graph-level score $S_r\in(0,1)$ measuring the overall plausibility of the plan under request r; node-level risk $\mathbf{p}^V_r=\{P^V_r(v)\}_{v\in V_r} $ indicating the probability that a particular step selects an incorrect tool; edge-level risk $\mathbf{p}^E_r=\{P^E_r(u,v)\}_{(u,v)\in E_r}$ indicating the probability that two adjacent steps are unreliably connected (e.g., missing intermediate steps). Concretely, we compute logits with three prediction heads $f_{\mathrm{g}}, f_{\mathrm{v}}, f_{\mathrm{e}}$ (lightweight MLPs) and map them to
probabilities via the sigmoid function $\sigma(\cdot)$:
\begin{equation}
\label{eq:verifier-heads}
\begin{aligned}
z(G_r) &= f_{\mathrm{g}}\!\big(\mathbf{h}_G\big),
& S_r &= \sigma\!\big(z(G_r)\big),\\
z_r^V(v) &= f_{\mathrm{v}}\!\big(\mathbf{h}_v\big),
& P_r^V(v) &= \sigma\!\big(z_r^V(v)\big),\\
z_r^E(u,v) &= f_{\mathrm{e}}\!\Big([\mathbf{h}_u;\ \mathbf{h}_v;\ \mathbf{x}_{uv}]\Big),
& P_r^E(u,v) &= \sigma\!\big(z_r^E(u,v)\big).
\end{aligned}
\end{equation}

\subsection{Perturbation-based Supervision}
In this subsection, we perturb the ground truth plan graphs in the training data to generate supervision signals.
\subsubsection{Perturbation Operators}
\label{sec:perturb-ops}
Since real planning data typically lacks fine-grained annotations for incorrect plans and error locations,
we construct controllable perturbations from the ground truth plan graph $G^{\mathrm{gt}}_r$ to obtain perturbed graphs
$G^{\mathrm{pert}}_r$ with an operation log $\mathrm{ops}(G^{\mathrm{pert}}_r)$.
We consider two common failure modes.

\textbf{Wrong Tool.}
We select a node $v_i=(t_i,s_i)$ and \textsc{REPLACE} its tool with $t_i'$.
The substitute is sampled preferentially from the similar tool neighborhood $\mathcal{N}_K(t_i)$ to form hard negatives.
To avoid trivially infeasible samples, we require the replaced tool to satisfy interface connectivity with adjacent nodes
(otherwise resample), while keeping $s_i$ unchanged.

\textbf{Missing Step.}
We simulate missing intermediate steps and over-simplification by operating on a consecutive directed subchain
$t_u\rightarrow t_{u+1}\rightarrow\cdots\rightarrow t_v$ in $G^{gt}_r$, using two forms:
\textsc{DROP(span)} deletes at least one intermediate node and directly connects the endpoint nodes to form $t_u\rightarrow t_v$. We require the newly created shortcut edge(s) to satisfy interface type connectivity; otherwise we resample the span, so that the perturbed plan remains type-executable while
being semantically missing steps.
\textsc{COMPRESS(span$\rightarrow$1)} replaces the intermediate nodes with a single tool node $t^*$, yielding $t_u\rightarrow t^*\rightarrow t_v$, where $t^*$ represents an over-generalized or improper merge of the original span. To keep the perturbation executable, we require both endpoint edges
to satisfy interface connectivity; otherwise we resample $t^*$ or reselect the span.
When sampling spans, we leverage tool sequence statistics $f_n(\cdot)$ to prioritize locally plausible paths, improving the
realism and difficulty of perturbations.

The two operators can be applied independently or composed multiple times on the same
$G^{gt}_r$, yielding candidates with different severities and error locations. Detailed perturbation specifications are provided in Appendix~\ref{app:perturbation_operators}.

\subsubsection{Supervision Signals}
\label{sec:supervision}
We derive graph-level soft targets and local targets automatically from perturbation logs.

\textbf{Graph-level target.}
For each perturbed graph $G^{\mathrm{pert}}_r$, we define a non-negative perturbation cost
$c(G^{\mathrm{pert}}_r)=\sum_{o\in \mathrm{ops}(G^{\mathrm{pert}}_r)}\eta(o)$, 
where $\eta(o)$ is an operation-type dependent penalty that reflects the strength of the applied perturbation, larger cost indicates more severe corruption. The detailed computation of $\eta(o)$ is provided in Appendix~\ref{app:cost}. We then map the cost to a graph-level soft target:
\begin{equation}
\label{eq:soft-target}
y(G^{\mathrm{pert}}_r)=\exp\!\left(-\frac{c(G^{\mathrm{pert}}_r)}{\tau}\right)\in(0,1],
\end{equation}
where $\tau$ is a hyperparameter. For the ground truth plan graph $G^{\mathrm{gt}}_r$, we set $c(G^{\mathrm{gt}}_r)=0$ and thus $y(G^{\mathrm{gt}}_r)=1$.

\textbf{Node-level target.}
We set $\ell^{\mathrm{node}}_v=1$ if node $v$ is replaced by \textsc{REPLACE} or introduced as the compressed node $t^*$
in \textsc{COMPRESS}; otherwise $\ell^{\mathrm{node}}_v=0$.

\textbf{Edge-level target.}
We set $\ell^{\mathrm{edge}}_{uv}=1$ for shortcut edges created by \textsc{DROP} and edges created by
\textsc{COMPRESS}; otherwise $\ell^{\mathrm{edge}}_{uv}=0$.

\subsubsection{Two-stage Training}
\label{sec:training}
We adopt a two-stage training strategy: Stage~I learns graph-level scoring, and Stage~II learns fine-grained local diagnosis. The detailed training losses are described in Appendix~\ref{app:two_stage_training}.

\textbf{Stage I: Graph-level training.} In the first stage, we train the graph encoder and graph-level prediction head to produce a continuous score for overall plan quality. Since we have a set of perturbed graphs generated for the same request $r$, we employ a margin-based ranking loss $\mathcal{L}_{\mathrm{rank}}$. We further integrate it with the typical binary cross-entropy loss $\mathcal{L}_{\mathrm{graph}}$ as $\mathcal{L}_{\mathrm{stage1}}=\mathcal{L}_{\mathrm{rank}}+\lambda_{\mathrm{graph}}\mathcal{L}_{\mathrm{graph}}.$

\textbf{Stage II: Node/Edge-level training.}
In Stage~II, we freeze most encoder parameters and fine-tune only the last GNN layer, as well as node/edge-level prediction heads. Here we combine the binary cross-entropy losses for nodes and edges as $\mathcal{L}_{\mathrm{stage2}}=\mathcal{L}_{\mathrm{node}}+\lambda_{\mathrm{edge}}\mathcal{L}_{\mathrm{edge}}.$

\subsection{Verification-guided Local Correction}
\label{sec:inference-local-correction}

At inference time, we will improve the original plan based on the graph-level score $S_r$, node-level risks $\mathbf{p}^V_r$ and edge-level risks $\mathbf{p}^E_r$.

\subsubsection{Decision and Editable Region}
We will modify the original plan based on local correction if and only if $S_r < \tau_G$. When local correction is triggered, we threshold the continuous risks to obtain editable regions:
\begin{equation}
\begin{aligned}
\mathcal{V}_{\text{edit}} &= \{v\in V_r:\ P^V_r(v)\ge \tau_V\},\\
\mathcal{E}_{\text{edit}} &= \{(u,v)\in E_r:\ P^E_r(u,v)\ge \tau_E\},
\end{aligned}
\label{eq:editable-region}
\end{equation}
where $\tau_G,\tau_V,\tau_E\in(0,1)$ are the graph-level acceptance threshold, node-risk threshold, and edge-risk threshold, respectively.
$\mathcal{V}_{\text{edit}}$ indicates positions that may contain wrong tool selections, while $\mathcal{E}_{\text{edit}}$ indicates potentially unreliable transitions (e.g., missing intermediate steps or excessive compression).

\subsubsection{LLM-guided Correction}
We design a constrained editing prompt and feed it back to the LLM planner for correction. 
The prompt includes $r$, $G_r$, $S_r$, the editable sets $\mathcal{V}_{\text{edit}}$ and $\mathcal{E}_{\text{edit}}$, the local risks $\mathbf{p}^V_r$ and $\mathbf{p}^E_r$, as well as the candidate tools $\{\mathcal{C}^{\text{rep}}(v)\}$ for node replacement and $\{\mathcal{C}^{\text{ins}}(u,v)\}$ for edge insertion. We present the details about candidate tools in Appendix~\ref{app:candidate_tools}. The LLM output is restricted to a sequence of at most $K_{\max}$ edits, where each edit must be chosen from a predefined set of operations:
\begin{itemize}
  \item \texttt{replace\_on\_node(node\_id, candidate\_id, step)}: for a high-risk node $v_i\in\mathcal{V}_{\mathrm{edit}}$, select $t\in\mathcal{C}^{\mathrm{rep}}(v_i)$ to replace the tool and generate the updated step text \texttt{step};
  \item \texttt{insert\_on\_edge(edge\_id, candidate\_id, step)}: for a high-risk edge $(u,v)\in\mathcal{E}_{\mathrm{edit}}$, select $t\in\mathcal{C}^{\mathrm{ins}}(u,v)$ to insert a new node between the endpoints, rewire the graph accordingly, and generate the inserted step text \texttt{step};
  \item \texttt{no\_change()}: perform no modification.
\end{itemize}

Then we apply the edits accordingly to obtain a corrected plan $G'_r$. We will accept $G'_r$ if its graph-level score is higher than the original $G_r$. Otherwise, we will keep the plan unchanged.

\begin{table*}[t]
\caption{Performance comparison across four datasets on GPT-4o: Node-F1, Link-F1, and Accuracy are
reported in \%. The best results are highlighted in boldface, and the second-best results are underlined.}
\label{tab:main_results}
\centering
\resizebox{1.0\textwidth}{!}{
\setlength{\tabcolsep}{5pt}
\renewcommand{\arraystretch}{1.25}
\begin{tabular}{p{7em}l ccc ccc ccc ccc}
\toprule
\multicolumn{1}{p{7em}}{\multirow{2}{*}{}} & \multirow{2}{*}{\textbf{Method}}
& \multicolumn{3}{c}{\textbf{HuggingFace}}
& \multicolumn{3}{c}{\textbf{Multimedia}}
& \multicolumn{3}{c}{\textbf{DailyLife}}
& \multicolumn{3}{c}{\textbf{UltraTool}} \\
\cmidrule(lr){3-5} \cmidrule(lr){6-8} \cmidrule(lr){9-11} \cmidrule(lr){12-14}
& & \textit{n-F1}$\uparrow$ & \textit{l-F1}$\uparrow$ & \textit{Acc}$\uparrow$
  & \textit{n-F1}$\uparrow$ & \textit{l-F1}$\uparrow$ & \textit{Acc}$\uparrow$
  & \textit{n-F1}$\uparrow$ & \textit{l-F1}$\uparrow$ & \textit{Acc}$\uparrow$
  & \textit{n-F1}$\uparrow$ & \textit{l-F1}$\uparrow$ & \textit{Acc}$\uparrow$ \\
\midrule

\multicolumn{1}{c}{\multirow{5}{*}{\textbf{Direct}}}
& \cellcolor{gray!15} Raw & \cellcolor{gray!15} 79.60 & \cellcolor{gray!15} 55.27 & \cellcolor{gray!15} 34.20 & \cellcolor{gray!15} 85.94 & \cellcolor{gray!15} 64.36 & \cellcolor{gray!15} 49.60 & \cellcolor{gray!15} 97.12 & \cellcolor{gray!15} 84.21 & \cellcolor{gray!15} \underline{72.80} & \cellcolor{gray!15} 73.07 & \cellcolor{gray!15} 46.36 & \cellcolor{gray!15} 36.80 \\
& \hspace*{2pt}+Refine & \hspace*{2pt}78.39 & \hspace*{2pt}52.86 & \hspace*{2pt}31.20 & \hspace*{2pt}\underline{87.02} & \hspace*{2pt}65.98 & \hspace*{2pt}48.40 & \hspace*{2pt}96.93 & \hspace*{2pt}60.12 & \hspace*{2pt}48.20 & \hspace*{2pt}73.30 & \hspace*{2pt}38.92 & \hspace*{2pt}30.80 \\
& \cellcolor{gray!15} +VeriCoder & \cellcolor{gray!15} \underline{79.78} & \cellcolor{gray!15} 55.45 & \cellcolor{gray!15} 34.00 & \cellcolor{gray!15} 86.69 & \cellcolor{gray!15} \underline{66.17} & \cellcolor{gray!15} 50.70 & \cellcolor{gray!15} \underline{97.44} & \cellcolor{gray!15} 56.81 & \cellcolor{gray!15} 46.80 & \cellcolor{gray!15} 73.86 & \cellcolor{gray!15} 34.11 & \cellcolor{gray!15} 26.45 \\
& \hspace*{2pt}+VeriPlan & \hspace*{2pt}79.64 & \hspace*{2pt}\underline{55.95} & \hspace*{2pt}\underline{34.80} & \hspace*{2pt}86.67 & \hspace*{2pt}66.14 & \hspace*{2pt}\underline{51.60} & \hspace*{2pt}97.12 & \hspace*{2pt}\underline{84.22} & \hspace*{2pt}\underline{72.80} & \hspace*{2pt}\textbf{78.26} & \hspace*{2pt}\textbf{53.89} & \hspace*{2pt}\underline{42.40} \\
& \cellcolor{gray!15} \textbf{+GNNVerifier} & \cellcolor{gray!15} \textbf{82.82} & \cellcolor{gray!15} \textbf{60.71} & \cellcolor{gray!15} \textbf{43.80} & \cellcolor{gray!15} \textbf{88.09} & \cellcolor{gray!15} \textbf{70.74} & \cellcolor{gray!15} \textbf{58.60} & \cellcolor{gray!15} \textbf{97.51} & \cellcolor{gray!15} \textbf{87.45} & \cellcolor{gray!15} \textbf{78.76} & \cellcolor{gray!15} \underline{76.89} & \cellcolor{gray!15} \underline{52.82} & \cellcolor{gray!15} \textbf{42.80} \\
\midrule

\multicolumn{1}{c}{\multirow{5}{*}{\textbf{ReAct}}}
& \cellcolor{gray!15} Raw & \cellcolor{gray!15} 79.91 & \cellcolor{gray!15} 53.35 & \cellcolor{gray!15} 31.80 & \cellcolor{gray!15} 86.50 & \cellcolor{gray!15} 64.53 & \cellcolor{gray!15} 48.80 & \cellcolor{gray!15} 96.59 & \cellcolor{gray!15} 57.01 & \cellcolor{gray!15} \underline{45.20} & \cellcolor{gray!15} 73.89 & \cellcolor{gray!15} 39.35 & \cellcolor{gray!15} 32.40 \\
& \hspace*{2pt}+Refine & \hspace*{2pt}78.19 & \hspace*{2pt}52.81 & \hspace*{2pt}30.80 & \hspace*{2pt}86.61 & \hspace*{2pt}65.60 & \hspace*{2pt}47.80 & \hspace*{2pt}96.25 & \hspace*{2pt}51.89 & \hspace*{2pt}38.08 & \hspace*{2pt}74.59 & \hspace*{2pt}36.34 & \hspace*{2pt}29.80 \\
& \cellcolor{gray!15} +VeriCoder & \cellcolor{gray!15} \underline{80.94} & \cellcolor{gray!15} \underline{58.31} & \cellcolor{gray!15} \underline{38.20} & \cellcolor{gray!15} \underline{90.37} & \cellcolor{gray!15} \underline{73.67} & \cellcolor{gray!15} \underline{55.80} & \cellcolor{gray!15} \textbf{97.34} & \cellcolor{gray!15} 36.09 & \cellcolor{gray!15} 20.08 & \cellcolor{gray!15} 73.43 & \cellcolor{gray!15} 28.51 & \cellcolor{gray!15} 21.44 \\
& \hspace*{2pt}+VeriPlan & \hspace*{2pt}80.16 & \hspace*{2pt}54.04 & \hspace*{2pt}33.40 & \hspace*{2pt}87.45 & \hspace*{2pt}68.52 & \hspace*{2pt}52.00 & \hspace*{2pt}\underline{96.62} & \hspace*{2pt}\underline{57.15} & \hspace*{2pt}\underline{45.20} & \hspace*{2pt}\textbf{77.91} & \hspace*{2pt}\underline{46.31} & \hspace*{2pt}\underline{34.47} \\
& \cellcolor{gray!15} \textbf{+GNNVerifier} & \cellcolor{gray!15} \textbf{82.40} & \cellcolor{gray!15} \textbf{60.66} & \cellcolor{gray!15} \textbf{43.60} & \cellcolor{gray!15} \textbf{90.46} & \cellcolor{gray!15} \textbf{73.73} & \cellcolor{gray!15} \textbf{59.00} & \cellcolor{gray!15} 96.59 & \cellcolor{gray!15} \textbf{85.83} & \cellcolor{gray!15} \textbf{76.35} & \cellcolor{gray!15} \textbf{77.91} & \cellcolor{gray!15} \textbf{52.06} & \cellcolor{gray!15} \textbf{44.20} \\
\midrule

\multicolumn{1}{c}{\multirow{5}{*}{\textbf{GNN4Plan}}}
& \cellcolor{gray!15} Raw & \cellcolor{gray!15} 78.68 & \cellcolor{gray!15} \underline{57.99} & \cellcolor{gray!15} \underline{41.20} & \cellcolor{gray!15} 84.91 & \cellcolor{gray!15} 69.41 & \cellcolor{gray!15} \underline{57.00} & \cellcolor{gray!15} \underline{97.25} & \cellcolor{gray!15} \underline{87.51} & \cellcolor{gray!15} \underline{78.80} & \cellcolor{gray!15} 71.68 & \cellcolor{gray!15} \underline{46.99} & \cellcolor{gray!15} \underline{37.80} \\
& \hspace*{2pt}+Refine & \hspace*{2pt}77.52 & \hspace*{2pt}51.92 & \hspace*{2pt}32.40 & \hspace*{2pt}87.64 & \hspace*{2pt}69.82 & \hspace*{2pt}54.20 & \hspace*{2pt}96.63 & \hspace*{2pt}53.46 & \hspace*{2pt}43.09 & \hspace*{2pt}\underline{75.60} & \hspace*{2pt}42.59 & \hspace*{2pt}34.61 \\
& \cellcolor{gray!15} +VeriCoder & \cellcolor{gray!15} \underline{80.01} & \cellcolor{gray!15} 56.33 & \cellcolor{gray!15} 38.55 & \cellcolor{gray!15} \underline{87.79} & \cellcolor{gray!15} \underline{70.40} & \cellcolor{gray!15} 56.20 & \cellcolor{gray!15} \textbf{97.65} & \cellcolor{gray!15} 50.37 & \cellcolor{gray!15} 42.48 & \cellcolor{gray!15} 75.17 & \cellcolor{gray!15} 36.31 & \cellcolor{gray!15} 27.25 \\
& \hspace*{2pt}+VeriPlan & \hspace*{2pt}78.68 & \hspace*{2pt}\underline{57.99} & \hspace*{2pt}\underline{41.20} & \hspace*{2pt}84.91 & \hspace*{2pt}69.41 & \hspace*{2pt}\underline{57.00} & \hspace*{2pt}\underline{97.25} & \hspace*{2pt}\underline{87.51} & \hspace*{2pt}\underline{78.80} & \hspace*{2pt}71.68 & \hspace*{2pt}\underline{46.99} & \hspace*{2pt}\underline{37.80} \\
& \cellcolor{gray!15} \textbf{+GNNVerifier} & \cellcolor{gray!15} \textbf{82.69} & \cellcolor{gray!15} \textbf{60.79} & \cellcolor{gray!15} \textbf{43.80} & \cellcolor{gray!15} \textbf{88.96} & \cellcolor{gray!15} \textbf{71.24} & \cellcolor{gray!15} \textbf{60.60} & \cellcolor{gray!15} \underline{97.25} & \cellcolor{gray!15} \textbf{87.61} & \cellcolor{gray!15} \textbf{78.98} & \cellcolor{gray!15} \textbf{76.44} & \cellcolor{gray!15} \textbf{53.46} & \cellcolor{gray!15} \textbf{43.80} \\
\bottomrule
\end{tabular}
}
\end{table*}

\section{Experiments}
We conduct extensive experiments to answer the following research questions (RQs):
\textbf{RQ1:} How does the proposed GNNVerifier improve plan quality compared to state-of-the-art baselines across diverse datasets and planners?
\textbf{RQ2:} How do the key designs contribute individually to the overall effectiveness?
\textbf{RQ3:} How does the verification-guided correction reduce each type of plan error across different datasets?
\textbf{RQ4:} How do the learned graph-, node-, and edge-level embedding spaces separate incorrect samples from correct ones? 
Besides, we present two additional experiments in the appendix: hyperparameter analysis (Appendix~\ref{sec:hyperparam_analysis}) and case study (Appendix~\ref{app:case_study}).

\subsection{Experimental Setup}

\textbf{Datasets.}
We evaluate on two task-planning benchmarks:   \textbf{Task-Bench}~\cite{shen2024taskbench} and \textbf{UltraTool}~\cite{huang2024planning}.
TaskBench contains three datasets that require multi-step tool invocations:
\textbf{HuggingFace}, which focuses on planning for AI/ML workflows over HuggingFace models (e.g., model retrieval, selection, and pipeline composition);
\textbf{Multimedia}, which covers user-centric multimedia tasks such as file downloading, editing, and format conversion;
and \textbf{DailyLife}, which includes everyday service APIs such as web search, shopping, and information retrieval.
To assess scalability on larger dependency graphs, we additionally conduct the same experiments on UltraTool. All datasets provide a user request together with a ground truth plan graph (tool nodes and dependency links).
We follow the data preprocessing and format conventions in prior work (e.g., GNN4Plan~\cite{wu2024can}) to ensure fair comparison. Detailed dataset descriptions are provided in the Appendix~\ref{app:datasets}.

\textbf{Evaluation.}
For both benchmarks, we follow the public split used in GNN4Plan~\cite{wu2024can}: 3000 instances for training and 500 instances for testing on each dataset.
We further hold out 10\% of the training set as a validation split for hyperparameter and threshold selection.
We adopt the standard metrics from previous work~\cite{wu2024can,shen2024taskbench}:
\textbf{Node F1-score} (n-F1), which measures the accuracy of invoked tasks (tool nodes);
\textbf{Link F1-score} (l-F1), which measures the accuracy of invoked dependencies (directed links);
and \textbf{Accuracy} (Acc), which measures the task-level success rate by checking whether the predicted tasks and dependencies exactly match the ground truth. All experiments are conducted three times, and the results are reported as the average value.

\textbf{Backbone LLMs.}
We report the main results with two backbone LLMs, GPT-4o~\cite{achiam2023gpt} and Qwen3-235B-A22B-Instruct-2507~\cite{yang2025qwen3}. To ensure a fair comparison, within each experimental setting we use the same backbone LLM consistently across all modules, including the planner and the correction module. All additional experiments and analyses are conducted under GPT-4o.

\textbf{Planners.}
We apply our verifier to three planning methods:
\textbf{Direct}~\cite{shen2024taskbench}, LLM generates a complete plan in a single shot given the user request and tool descriptions;
\textbf{ReAct}~\cite{yao2022react}, LLM alternates between reasoning and acting to incrementally construct the plan;
and \textbf{GNN4Plan}~\cite{wu2024can}, which uses a GNN to guide plan construction by selecting tool nodes and tool-invocation edges.

\textbf{Baselines.}
On top of each planner, we compare several plan-correction strategies: 
\textbf{Refine}~\cite{madaan2023self}, an LLM-only self-refinement that revises the initial plan without explicit diagnostic signals;
\textbf{VeriCoder}~\cite{ho2025verilogcoder} employs an LLM-based plan verification assistant to check whether the decomposed sub-tasks are consistent with the user specification and to provide actionable suggestions when inconsistencies are found.
\textbf{VeriPlan}~\cite{lee2025veriplan} couples an LLM with an external verifier that checks whether a plan satisfies explicit constraints and uses the detected violations to guide a correction step. In our implementation, the verifier checks edges against the dependency graph specification and prompts the LLM to correct invalid dependencies. For a controlled comparison, all verifiers perform exactly one revision step.

\textbf{Implementation Details.}
All experiments are conducted on $8\times$ NVIDIA A800-40G GPUs.
We use E5-large~\cite{wang2022text} as the text encoder to obtain unified semantic representations for the user request, step texts, and tool descriptions.
For each tool $t$, we precompute its semantic neighborhood $\mathcal{N}_K(t)$ by cosine similarity with $K=10$.
Our verifier uses a 3-layer GNN. The GNN and all MLPs are set to a dimension of $1024$, with ReLU activations and a dropout rate of $0.1$. Specifically, the feature concatenations (node/edge) and step--tool alignment module are implemented as single-layer MLPs, while GNN message/update functions $\phi_{\mathrm{in/out}}^{(\ell)}$ and the update MLP in Eq.~\eqref{eq:msg-update} use two-layer MLPs. For the step--tool alignment module, early stopping is configured with a patience of 5 based on the training loss.
The graph representation is obtained by mean pooling. We train the GNN verifier with learning rate $2\times 10^{-5}$ and batch size $512$. The correction specific thresholds are provided in Appendix~\ref{app:threshold_selection}, while hyperparameters related to loss weighting and soft targets are tuned on the validation set and analyzed in Appendix~\ref{sec:hyperparam_analysis}.

\begin{table}[t]
\caption{Ablation studies on HuggingFace and Multimedia: Node-F1, Link-F1, and Accuracy are reported in \%.}
\label{tab:ablation_results}
\centering
\resizebox{\columnwidth}{!}{
\setlength{\tabcolsep}{4pt}
\renewcommand{\arraystretch}{1.2}
\begin{tabular}{c l ccc ccc}
\toprule
& \multirow{2}{*}{\textbf{Variant}}
& \multicolumn{3}{c}{\textbf{HuggingFace}}
& \multicolumn{3}{c}{\textbf{Multimedia}} \\
\cmidrule(lr){3-5} \cmidrule(lr){6-8}
& & \textit{n-F1}$\uparrow$ & \textit{l-F1}$\uparrow$ & \textit{Acc}$\uparrow$
  & \textit{n-F1}$\uparrow$ & \textit{l-F1}$\uparrow$ & \textit{Acc}$\uparrow$ \\
\midrule

\multirow{10}{*}{\rotatebox{90}{\textbf{Direct}}}
& \hspace*{2pt}Raw & \hspace*{2pt}79.60 & \hspace*{2pt}55.27 & \hspace*{2pt}34.20 & \hspace*{2pt}85.94 & \hspace*{2pt}64.36 & \hspace*{2pt}49.60 \\
& \cellcolor{gray!15} w/o GNN & \cellcolor{gray!15} 75.78 & \cellcolor{gray!15} 51.77 & \cellcolor{gray!15} 33.00 & \cellcolor{gray!15} 84.40 & \cellcolor{gray!15} 66.35 & \cellcolor{gray!15} 50.20 \\
& \hspace*{2pt}w/o Stage-II & \hspace*{2pt}79.92 & \hspace*{2pt}55.51 & \hspace*{2pt}35.40 & \hspace*{2pt}86.03 & \hspace*{2pt}66.05 & \hspace*{2pt}50.40 \\
& \cellcolor{gray!15} w/o Stage-I & \cellcolor{gray!15} 79.60 & \cellcolor{gray!15} 57.86 & \cellcolor{gray!15} \underline{39.80} & \cellcolor{gray!15} 85.94 & \cellcolor{gray!15} \underline{68.97} & \cellcolor{gray!15} \underline{56.40} \\
& \hspace*{2pt}w/o Node Feat. & \hspace*{2pt}80.36 & \hspace*{2pt}57.94 & \hspace*{2pt}35.60 & \hspace*{2pt}86.01 & \hspace*{2pt}64.53 & \hspace*{2pt}50.00 \\
& \cellcolor{gray!15} w/o Edge Feat. & \cellcolor{gray!15} \underline{80.53} & \cellcolor{gray!15} 56.48 & \cellcolor{gray!15} 37.20 & \cellcolor{gray!15} \underline{86.44} & \cellcolor{gray!15} 66.61 & \cellcolor{gray!15} 51.80 \\
& \hspace*{2pt}w/o Graph FB & \hspace*{2pt}80.19 & \hspace*{2pt}55.97 & \hspace*{2pt}35.60 & \hspace*{2pt}84.14 & \hspace*{2pt}64.47 & \hspace*{2pt}49.00 \\
& \cellcolor{gray!15} w/o Node FB & \cellcolor{gray!15} 80.45 & \cellcolor{gray!15} \underline{58.06} & \cellcolor{gray!15} 34.80 & \cellcolor{gray!15} 86.03 & \cellcolor{gray!15} 64.70 & \cellcolor{gray!15} 50.00 \\
& \hspace*{2pt}w/o Edge FB & \hspace*{2pt}80.36 & \hspace*{2pt}56.21 & \hspace*{2pt}34.80 & \hspace*{2pt}86.30 & \hspace*{2pt}67.59 & \hspace*{2pt}52.20 \\
& \cellcolor{gray!15} Full & \cellcolor{gray!15} \textbf{82.82} & \cellcolor{gray!15} \textbf{60.71} & \cellcolor{gray!15} \textbf{43.80} & \cellcolor{gray!15} \textbf{88.09} & \cellcolor{gray!15} \textbf{70.74} & \cellcolor{gray!15} \textbf{58.60} \\
\midrule

\multirow{10}{*}{\rotatebox{90}{\textbf{ReAct}}}
& \hspace*{2pt}Raw & \hspace*{2pt}79.91 & \hspace*{2pt}53.35 & \hspace*{2pt}31.80 & \hspace*{2pt}86.50 & \hspace*{2pt}64.53 & \hspace*{2pt}48.80 \\
& \cellcolor{gray!15} w/o GNN & \cellcolor{gray!15} 75.58 & \cellcolor{gray!15} 52.59 & \cellcolor{gray!15} 31.00 & \cellcolor{gray!15} 84.92 & \cellcolor{gray!15} 65.38 & \cellcolor{gray!15} 46.80 \\
& \hspace*{2pt}w/o Stage-II & \hspace*{2pt}80.21 & \hspace*{2pt}54.12 & \hspace*{2pt}35.20 & \hspace*{2pt}87.73 & \hspace*{2pt}65.28 & \hspace*{2pt}48.80 \\
& \cellcolor{gray!15} w/o Stage-I & \cellcolor{gray!15} 79.91 & \cellcolor{gray!15} \underline{56.50} & \cellcolor{gray!15} \underline{41.00} & \cellcolor{gray!15} 86.50 & \cellcolor{gray!15} \underline{69.25} & \cellcolor{gray!15} \underline{56.40} \\
& \hspace*{2pt}w/o Node Feat. & \hspace*{2pt}80.12 & \hspace*{2pt}54.40 & \hspace*{2pt}36.40 & \hspace*{2pt}87.15 & \hspace*{2pt}64.83 & \hspace*{2pt}49.00 \\
& \cellcolor{gray!15} w/o Edge Feat. & \cellcolor{gray!15} \underline{81.23} & \cellcolor{gray!15} 54.04 & \cellcolor{gray!15} 37.60 & \cellcolor{gray!15} 87.61 & \cellcolor{gray!15} 67.36 & \cellcolor{gray!15} 51.00 \\
& \hspace*{2pt}w/o Graph FB & \hspace*{2pt}77.29 & \hspace*{2pt}51.24 & \hspace*{2pt}32.80 & \hspace*{2pt}84.87 & \hspace*{2pt}65.98 & \hspace*{2pt}49.20 \\
& \cellcolor{gray!15} w/o Node FB & \cellcolor{gray!15} 79.59 & \cellcolor{gray!15} 50.77 & \cellcolor{gray!15} 31.20 & \cellcolor{gray!15} 87.35 & \cellcolor{gray!15} 65.82 & \cellcolor{gray!15} 49.20 \\
& \hspace*{2pt}w/o Edge FB & \hspace*{2pt}79.56 & \hspace*{2pt}52.34 & \hspace*{2pt}30.80 & \hspace*{2pt}\underline{88.67} & \hspace*{2pt}67.42 & \hspace*{2pt}51.60 \\
& \cellcolor{gray!15} Full & \cellcolor{gray!15} \textbf{82.40} & \cellcolor{gray!15} \textbf{60.66} & \cellcolor{gray!15} \textbf{43.60} & \cellcolor{gray!15} \textbf{90.46} & \cellcolor{gray!15} \textbf{73.73} & \cellcolor{gray!15} \textbf{59.00} \\
\midrule

\multirow{10}{*}{\rotatebox{90}{\textbf{GNN4Plan}}}
& \hspace*{2pt}Raw & \hspace*{2pt}78.68 & \hspace*{2pt}57.99 & \hspace*{2pt}41.20 & \hspace*{2pt}84.91 & \hspace*{2pt}69.41 & \hspace*{2pt}57.00 \\
& \cellcolor{gray!15} w/o GNN & \cellcolor{gray!15} 76.91 & \cellcolor{gray!15} 52.70 & \cellcolor{gray!15} 33.00 & \cellcolor{gray!15} 83.39 & \cellcolor{gray!15} 66.66 & \cellcolor{gray!15} 50.00 \\
& \hspace*{2pt}w/o Stage-II & \hspace*{2pt}79.41 & \hspace*{2pt}55.26 & \hspace*{2pt}39.20 & \hspace*{2pt}85.66 & \hspace*{2pt}70.04 & \hspace*{2pt}\underline{58.40} \\
& \cellcolor{gray!15} w/o Stage-I & \cellcolor{gray!15} 79.60 & \cellcolor{gray!15} 57.86 & \cellcolor{gray!15} 39.80 & \cellcolor{gray!15} 85.91 & \cellcolor{gray!15} 69.41 & \cellcolor{gray!15} 57.00 \\
& \hspace*{2pt}w/o Node Feat. & \hspace*{2pt}79.22 & \hspace*{2pt}\underline{59.12} & \hspace*{2pt}42.00 & \hspace*{2pt}85.37 & \hspace*{2pt}70.03 & \hspace*{2pt}57.00 \\
& \cellcolor{gray!15} w/o Edge Feat. & \cellcolor{gray!15} \underline{80.70} & \cellcolor{gray!15} 58.38 & \cellcolor{gray!15} \underline{42.60} & \cellcolor{gray!15} 86.19 & \cellcolor{gray!15} 69.84 & \cellcolor{gray!15} 58.20 \\
& \hspace*{2pt}w/o Graph FB & \hspace*{2pt}77.07 & \hspace*{2pt}56.04 & \hspace*{2pt}41.20 & \hspace*{2pt}84.96 & \hspace*{2pt}69.45 & \hspace*{2pt}57.00 \\
& \cellcolor{gray!15} w/o Node FB & \cellcolor{gray!15} 80.33 & \cellcolor{gray!15} 59.01 & \cellcolor{gray!15} 42.00 & \cellcolor{gray!15} 84.96 & \cellcolor{gray!15} \underline{70.45} & \cellcolor{gray!15} 57.60 \\
& \hspace*{2pt}w/o Edge FB & \hspace*{2pt}80.33 & \hspace*{2pt}59.01 & \hspace*{2pt}42.00 & \hspace*{2pt}\underline{86.87} & \hspace*{2pt}69.41 & \hspace*{2pt}57.60 \\
& \cellcolor{gray!15} Full & \cellcolor{gray!15} \textbf{82.69} & \cellcolor{gray!15} \textbf{60.79} & \cellcolor{gray!15} \textbf{43.80} & \cellcolor{gray!15} \textbf{88.96} & \cellcolor{gray!15} \textbf{71.24} & \cellcolor{gray!15} \textbf{60.60} \\
\bottomrule
\end{tabular}
}
\end{table}

\subsection{Main Results (RQ1)}
Table~\ref{tab:main_results} reports the performance of different planner--verifier combinations on four datasets with GPT-4o. The counterpart results with Qwen3-235B-A22B-Instruct-2507 are provided in Appendix~\ref{app:main_results_qwen3}. Overall, our graph-based verifier improves node-level and edge-level planning metrics (n-F1 and l-F1) in most settings, and consistently achieves the best task-level accuracy.

\textbf{Broad generalization across datasets.}
Averaged over the three planners and compared to the best baseline, VeriPlan, our method achieves relative improvements in n-F1/l-F1/Acc of 3.95\% / 8.44\% /19.93\% on HuggingFace, 3.27\% / 5.70\% / 10.96\% on Multimedia, 0.12\% / 13.99\% / 18.95\% on DailyLife, and 1.49\% / 7.58\% / 14.07\% on UltraTool.
The improvements hold across datasets with substantially different tool inventories and task distributions, indicating strong generalization.
Notably, the n-F1 gain on DailyLife is small because the baseline node predictions are already near-saturated (n-F1 is typically above 96\% across planners in Table~\ref{tab:main_results}), leaving limited room for further improvement.

\textbf{Robust improvements across planners.}
Averaged over the four datasets and compared to the best baseline, VeriPlan, our method yields relative improvements in n-F1/l-F1/Acc of 1.06\% / 4.43\% / 11.09\% for Direct, 1.53\% / 20.47\% / 35.19\% for ReAct, and 3.86\% / 4.28\% / 5.76\% for GNN4Plan. The consistent gains across planners reflect improved robustness to heterogeneous error patterns produced by different planners.

\textbf{Stronger performance on the task-level metric.}
The n-F1 and l-F1 metrics quantify local matching quality for nodes and edges, whereas Acc is a task-level success rate that requires the predicted plan graph to be correct as a whole (both nodes and links). By modeling plans as attributed graphs and performing message passing over execution edges, our verifier better captures local context and structural consistency, leading to more reliable task accuracy across all datasets and planners.

\begin{figure*}[t]
  \centering
  \setlength{\tabcolsep}{6pt}
  \begin{tabular}{@{}c@{\hspace{8pt}}c@{\hspace{8pt}}c@{\hspace{8pt}}c@{}}
    \includegraphics[width=0.24\linewidth]{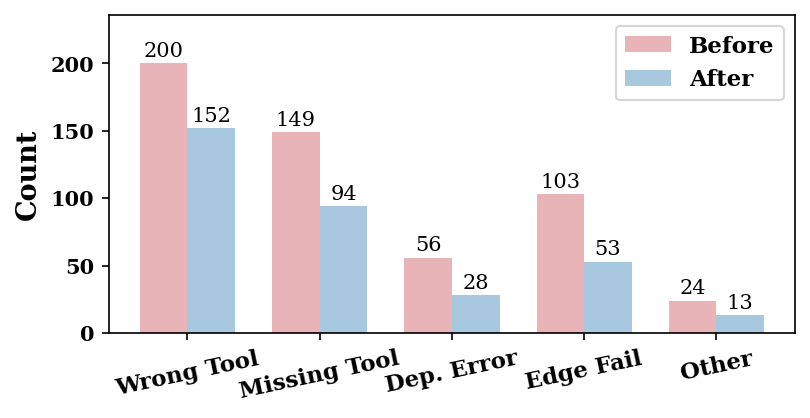} &
    \includegraphics[width=0.24\linewidth]{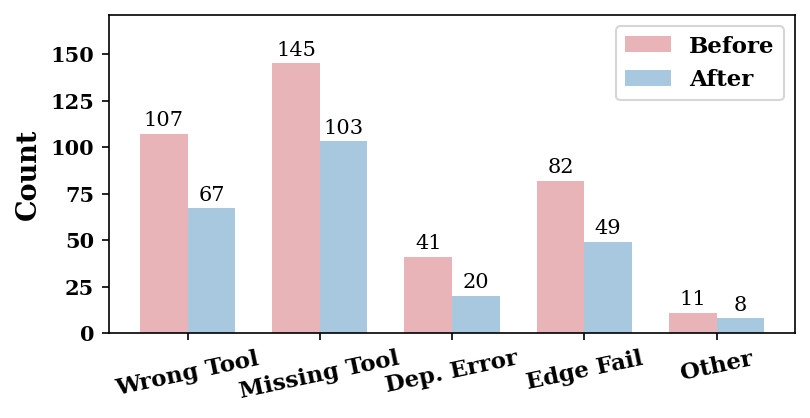} &
    \includegraphics[width=0.24\linewidth]{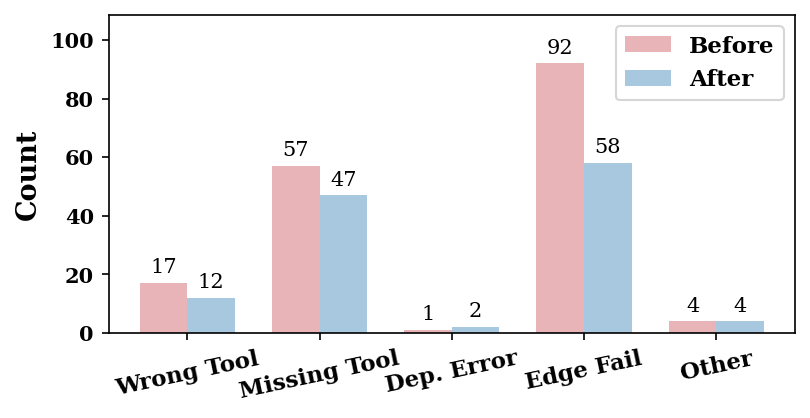} &
    \includegraphics[width=0.24\linewidth]{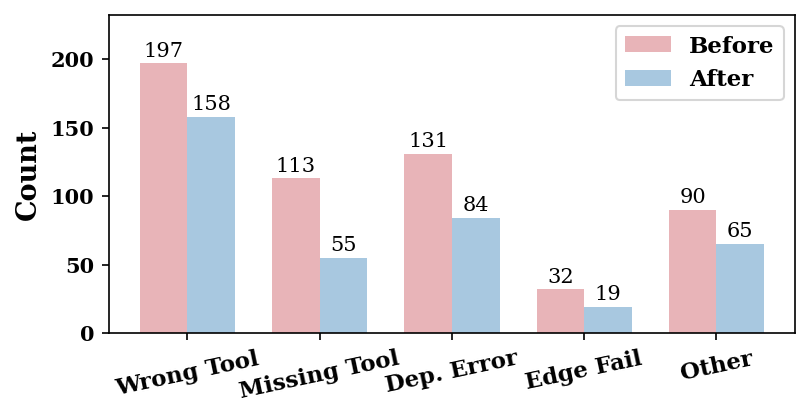}
    \\[0.4em]
    \small \textbf{(a)} \textbf{HuggingFace} & \small \textbf{(b)} \textbf{Multimedia} & \small \textbf{(c)} \textbf{DailyLife} & \small \textbf{(d)} \textbf{UltraTool}
  \end{tabular}
  \caption{Analysis of planning error types before and after correction across four datasets.}
  \label{fig:case_direct}
\end{figure*}

\begin{figure*}[t]
  \centering
  \setlength{\tabcolsep}{10pt} 
  \begin{subfigure}[t]{0.27\linewidth}
    \includegraphics[width=\linewidth]{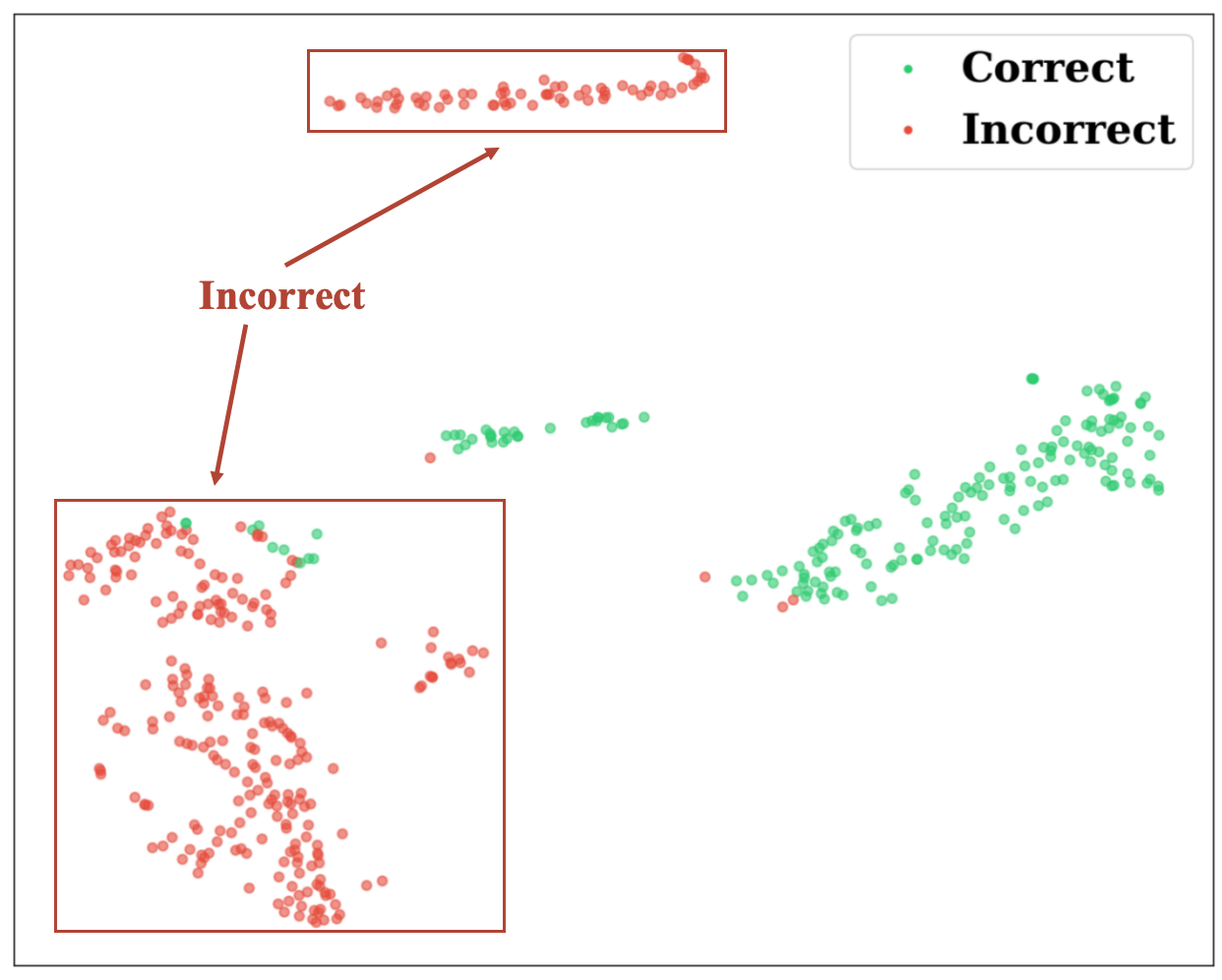}
    \caption{Graph}
    \label{subfig:ultratool_graph}
  \end{subfigure}
  \hspace{0.04\linewidth} 
  \begin{subfigure}[t]{0.27\linewidth}
    \centering
    \includegraphics[width=\linewidth]{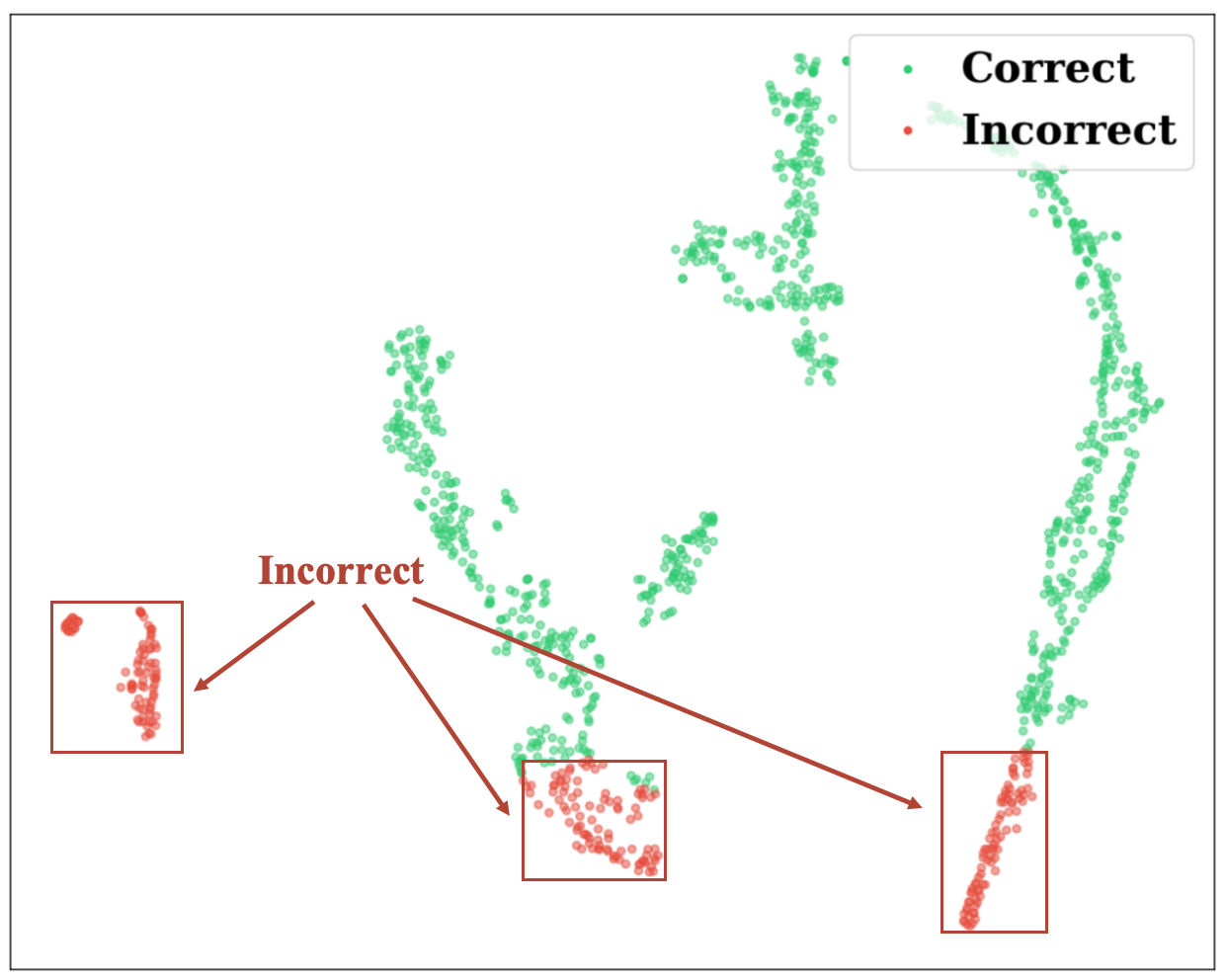}
    \caption{Node}
    \label{subfig:ultratool_node}
  \end{subfigure}
  \hspace{0.04\linewidth}
  \begin{subfigure}[t]{0.27\linewidth}
    \centering
    \includegraphics[width=\linewidth]{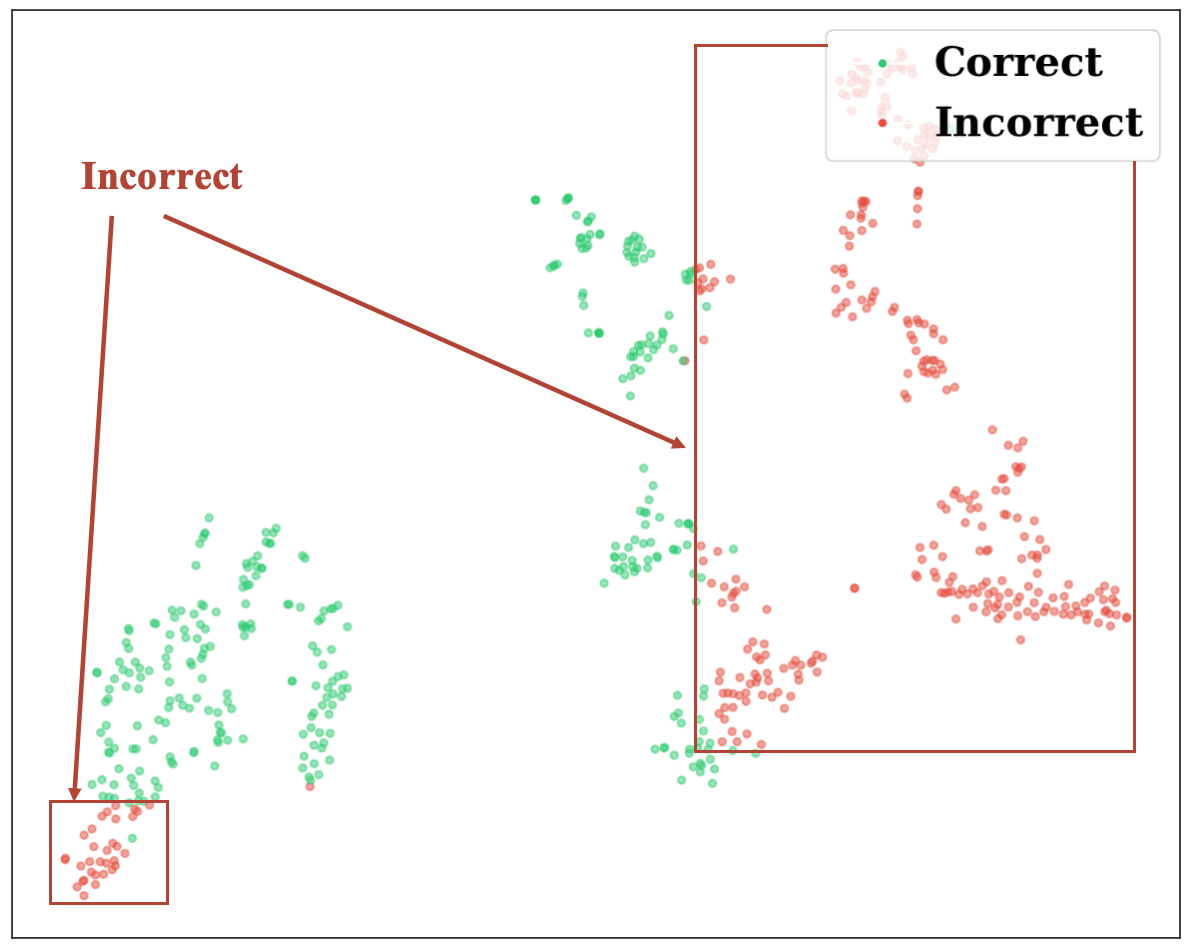}
    \caption{Edge}
    \label{subfig:ultratool_edge}
  \end{subfigure}

  \caption{The t-SNE visualization of graph-, node-, and edge-level embeddings on UltraTool.}
  \label{fig:ultratool_graph_node_edge}
\end{figure*}

\subsection{Ablation Studies (RQ2)}
We conduct ablation studies on all four datasets.
Table~\ref{tab:ablation_results} reports results on HuggingFace and Multimedia, while results on DailyLife and UltraTool are deferred to Appendix~\ref{app:abalation_study}. 
We compare our method with eight ablated variants to isolate the effects of (i) structural modeling and training objectives of the GNN verifier, (ii) the proposed advanced node/edge attributes, and (iii) the multi-granularity feedback used to guide LLM correction. 

\textbf{Necessity of graph-structured modeling.}
To verify that improvements come from structure-aware reasoning rather than merely richer features, we replace the GNN verifier with an MLP (w/o GNN), thereby removing message passing on the plan graph.
Notably, this ablation is not only substantially worse than the full model, but is also worse than having no verifier at all (Raw) in several settings.
For instance, under the GNN4Plan planner, Acc drops from 41.20\% (Raw) to 33.00\% on HuggingFace, from 57.00\% to 50.00\% on Multimedia.
This indicates that without relational modeling, verifier feedback can become harmful: an MLP tends to score nodes/edges largely independently and cannot aggregate contextual evidence along execution dependencies, making its prediction less precise and less globally consistent.

\textbf{Effectiveness of two-stage training.}
We ablate the two-stage training scheme by removing Stage-II (w/o Stage-II) or Stage-I (w/o Stage-I), testing whether graph-level quality learning and local error localization can substitute for each other.
The full model consistently performs best, indicating that the two stages provide complementary signals: Stage-I provides a global quality signal to assess overall plan soundness, while Stage-II provides precise local diagnostics that enable targeted and effective edits.

\textbf{Gains from advanced node/edge attributes.}
To verify our feature design, we keep only the basic features and ablate them separately: (i) w/o node feat., which removes step semantics and the step--tool matching features; and 
(ii) w/o edge feat., which removes the co-occurrence statistics. 
Results show that both ablations consistently underperform the full model across all planners on both Huggingface and Multimedia, suggesting a general benefit rather than a dataset specific artifact.

\textbf{Feedback signals for LLM-guided correction.}
We ablate the feedback provided to the LLM during local correction by removing:
graph-level scoring feedback (w/o graph FB), node-level risk feedback (w/o node FB), or edge-level risk feedback (w/o edge FB), together with their associated prompts, to test whether correction requires multi-granularity diagnostics rather than a single signal. Results show that the three feedback signals are complementary: graph feedback provides global calibration, node feedback supports tool replacement, and edge feedback supports dependency correction; using all of them yields the most stable improvements.

\subsection{Error Analysis (RQ3)}
\label{sec:error_type}
Figure~\ref{fig:case_direct} presents the distribution of five error types with the Direct planner, comparing the raw plans (\textit{Before}) and the plans after correction by our method (\textit{After}).
We define the error types as: (1) Wrong Tool (a step selects an incorrect tool), (2) Missing Tools (the plan omits one or more necessary intermediate steps), (3) Dependency Error (a predicted edge is I/O incompatible and thus does not exist in the dependency graph), (4) Edge Fail (the edge exists in the dependency graph but violates the user request), and (5) Other (e.g., empty outputs or redundant tools).

Across the four datasets, the raw plans are often coupled with local structural defects along execution dependencies. After applying our method, the error counts decrease broadly across types, with the most visible drops in Wrong Tool and Missing Tool, indicating that risk-guided local editing can effectively replace confused tools and insert missing intermediate steps. Dependency Error also reduces, reflecting that dependency graph type constraints provide a reliable guardrail during candidate generation and correction, while Edge Fail decreases as well, suggesting the verifier can detect and correct type compatible yet incorrect transitions by leveraging graph context rather than relying on interface compatibility alone. More detailed results are provided in the Appendix~\ref{app:error_breakdown_react_gnn4plan}.

\subsection{Visualization of GNN Embeddings (RQ4)}
Figure~\ref{fig:ultratool_graph_node_edge} visualizes the learned graph-, node-, and edge-level embeddings on the largest dataset UltraTool with the Direct planner. Overall, the t-SNE plots show a clear separation between correct and incorrect test samples with limited overlap across all three granularities.
At the graph level (Figure~\ref{fig:ultratool_graph_node_edge}(\subref{subfig:ultratool_graph})), correct plans form compact regions that are well separated from incorrect ones. This supports using the graph score for reliable acceptance versus correction decisions.
At the node and edge levels (Figure~\ref{fig:ultratool_graph_node_edge}(\subref{subfig:ultratool_node}) and (\subref{subfig:ultratool_edge})), nodes and edges from correct plans occupy more coherent clusters, while incorrect ones are scattered into different regions, suggesting that the verifier learns different patterns for identifying incorrect tool selections and transitions. The visualization suggests that our verifier learns consistently discriminative representations across graph, node, and edge granularities. More detailed visualization results and discussions are provided in Appendix~\ref{app:tsne_more}.

\section{Conclusion}
\label{app:conclusion}
In this paper, we propose an effective graph-based verifier for LLM task planning. By representing a generated plan as a directed graph with enriched attributes, we use a GNN to perform structural evaluation and diagnosis. We automatically generate training data for graph/node/edge-level scoring by perturbing ground truth plan graphs. Experimental results show that the feedback from our GNNVerifier can effectively assist an LLM to correct the original plans. For future work, a possible direction is to move from static verification to execution-aware verification and correction by integrating online signals from real tool calls.

\bibliographystyle{ACM-Reference-Format}
\bibliography{sample-base}

@String{Computing = "Computing" }

@String{Chelsea = "Chelsea" }

@ArtifactSoftware{R,
    title = {R: A Language and Environment for Statistical Computing},
    author = {{R Core Team}},
    organization = {R Foundation for Statistical Computing},
    address = {Vienna, Austria},
    year = {2019},
    url = {https://www.R-project.org/},
}

@article{wei2022chain,
  title={Chain-of-thought prompting elicits reasoning in large language models},
  author={Wei, Jason and Wang, Xuezhi and Schuurmans, Dale and Bosma, Maarten and Xia, Fei and Chi, Ed and Le, Quoc V and Zhou, Denny and others},
  journal={Advances in neural information processing systems},
  volume={35},
  pages={24824--24837},
  year={2022}
}

@inproceedings{yao2022react,
  title={React: Synergizing reasoning and acting in language models},
  author={Yao, Shunyu and Zhao, Jeffrey and Yu, Dian and Du, Nan and Shafran, Izhak and Narasimhan, Karthik R and Cao, Yuan},
  booktitle={The eleventh international conference on learning representations},
  year={2022}
}

@article{shen2023hugginggpt,
  title={Hugginggpt: Solving ai tasks with chatgpt and its friends in hugging face},
  author={Shen, Yongliang and Song, Kaitao and Tan, Xu and Li, Dongsheng and Lu, Weiming and Zhuang, Yueting},
  journal={Advances in Neural Information Processing Systems},
  volume={36},
  pages={38154--38180},
  year={2023}
}

@inproceedings{prasad2024adapt,
  title={Adapt: As-needed decomposition and planning with language models},
  author={Prasad, Archiki and Koller, Alexander and Hartmann, Mareike and Clark, Peter and Sabharwal, Ashish and Bansal, Mohit and Khot, Tushar},
  booktitle={Findings of the Association for Computational Linguistics: NAACL 2024},
  pages={4226--4252},
  year={2024}
}

@article{yao2023tree,
  title={Tree of thoughts: Deliberate problem solving with large language models, 2023},
  author={Yao, Shunyu and Yu, Dian and Zhao, Jeffrey and Shafran, Izhak and Griffiths, Thomas L and Cao, Yuan and Narasimhan, Karthik},
  journal={URL https://arxiv. org/abs/2305.10601},
  volume={3},
  pages={1},
  year={2023}
}

@inproceedings{besta2024graph,
  title={Graph of thoughts: Solving elaborate problems with large language models},
  author={Besta, Maciej and Blach, Nils and Kubicek, Ales and Gerstenberger, Robert and Podstawski, Michal and Gianinazzi, Lukas and Gajda, Joanna and Lehmann, Tomasz and Niewiadomski, Hubert and Nyczyk, Piotr and others},
  booktitle={Proceedings of the AAAI conference on artificial intelligence},
  volume={38},
  number={16},
  pages={17682--17690},
  year={2024}
}

@article{wang2022self,
  title={Self-consistency improves chain of thought reasoning in language models},
  author={Wang, Xuezhi and Wei, Jason and Schuurmans, Dale and Le, Quoc and Chi, Ed and Narang, Sharan and Chowdhery, Aakanksha and Zhou, Denny},
  journal={arXiv preprint arXiv:2203.11171},
  year={2022}
}

@article{liu2023llm+,
  title={Llm+ p: Empowering large language models with optimal planning proficiency},
  author={Liu, Bo and Jiang, Yuqian and Zhang, Xiaohan and Liu, Qiang and Zhang, Shiqi and Biswas, Joydeep and Stone, Peter},
  journal={arXiv preprint arXiv:2304.11477},
  year={2023}
}

@article{dagan2023dynamic,
  title={Dynamic planning with a llm},
  author={Dagan, Gautier and Keller, Frank and Lascarides, Alex},
  journal={arXiv preprint arXiv:2308.06391},
  year={2023}
}

@article{guan2023leveraging,
  title={Leveraging pre-trained large language models to construct and utilize world models for model-based task planning},
  author={Guan, Lin and Valmeekam, Karthik and Sreedharan, Sarath and Kambhampati, Subbarao},
  journal={Advances in Neural Information Processing Systems},
  volume={36},
  pages={79081--79094},
  year={2023}
}

@article{wu2024can,
  title={Can graph learning improve planning in LLM-based agents?},
  author={Wu, Xixi and Shen, Yifei and Shan, Caihua and Song, Kaitao and Wang, Siwei and Zhang, Bohang and Feng, Jiarui and Cheng, Hong and Chen, Wei and Xiong, Yun and others},
  journal={Advances in Neural Information Processing Systems},
  volume={37},
  pages={5338--5383},
  year={2024}
}

@article{cobbe2021training,
  title={Training verifiers to solve math word problems},
  author={Cobbe, Karl and Kosaraju, Vineet and Bavarian, Mohammad and Chen, Mark and Jun, Heewoo and Kaiser, Lukasz and Plappert, Matthias and Tworek, Jerry and Hilton, Jacob and Nakano, Reiichiro and others},
  journal={arXiv preprint arXiv:2110.14168},
  year={2021}
}

@inproceedings{lightman2023let,
  title={Let's verify step by step},
  author={Lightman, Hunter and Kosaraju, Vineet and Burda, Yuri and Edwards, Harrison and Baker, Bowen and Lee, Teddy and Leike, Jan and Schulman, John and Sutskever, Ilya and Cobbe, Karl},
  booktitle={The Twelfth International Conference on Learning Representations},
  year={2023}
}

@article{xiong2025rag,
  title={Rag-gym: Systematic optimization of language agents for retrieval-augmented generation},
  author={Xiong, Guangzhi and Jin, Qiao and Wang, Xiao and Fang, Yin and Liu, Haolin and Yang, Yifan and Chen, Fangyuan and Song, Zhixing and Wang, Dengyu and Zhang, Minjia and others},
  journal={arXiv preprint arXiv:2502.13957},
  year={2025}
}

@article{madaan2023self,
  title={Self-refine: Iterative refinement with self-feedback, 2023},
  author={Madaan, Aman and Tandon, Niket and Gupta, Prakhar and Hallinan, Skyler and Gao, Luyu and Wiegreffe, Sarah and Alon, Uri and Dziri, Nouha and Prabhumoye, Shrimai and Yang, Yiming and others},
  journal={URL https://arxiv. org/abs/2303.17651},
  year={2023}
}

@inproceedings{lee2025veriplan,
  title={Veriplan: Integrating formal verification and llms into end-user planning},
  author={Lee, Christine P and Porfirio, David and Wang, Xinyu Jessica and Zhao, Kevin Chenkai and Mutlu, Bilge},
  booktitle={Proceedings of the 2025 CHI Conference on Human Factors in Computing Systems},
  pages={1--19},
  year={2025}
}

@inproceedings{ho2025verilogcoder,
  title={Verilogcoder: Autonomous verilog coding agents with graph-based planning and abstract syntax tree (ast)-based waveform tracing tool},
  author={Ho, Chia-Tung and Ren, Haoxing and Khailany, Brucek},
  booktitle={Proceedings of the AAAI Conference on Artificial Intelligence},
  volume={39},
  number={1},
  pages={300--307},
  year={2025}
}

@inproceedings{kambhampati2024position,
  title={Position: LLMs can’t plan, but can help planning in LLM-modulo frameworks},
  author={Kambhampati, Subbarao and Valmeekam, Karthik and Guan, Lin and Verma, Mudit and Stechly, Kaya and Bhambri, Siddhant and Saldyt, Lucas Paul and Murthy, Anil B},
  booktitle={Forty-first International Conference on Machine Learning},
  year={2024}
}

@misc{anonymous2026unified,
  title={Unified Plan Verification with Static Rubrics and Dynamic Policies for Reliable {LLM} Planning},
  author={Anonymous},
  year={2026},
  url={https://openreview.net/forum?id=qDFegAnCin},
  note={Under review}
}

@article{shen2024taskbench,
  title={Taskbench: Benchmarking large language models for task automation},
  author={Shen, Yongliang and Song, Kaitao and Tan, Xu and Zhang, Wenqi and Ren, Kan and Yuan, Siyu and Lu, Weiming and Li, Dongsheng and Zhuang, Yueting},
  journal={Advances in Neural Information Processing Systems},
  volume={37},
  pages={4540--4574},
  year={2024}
}

@article{huang2024planning,
  title={Planning, creation, usage: Benchmarking llms for comprehensive tool utilization in real-world complex scenarios},
  author={Huang, Shijue and Zhong, Wanjun and Lu, Jianqiao and Zhu, Qi and Gao, Jiahui and Liu, Weiwen and Hou, Yutai and Zeng, Xingshan and Wang, Yasheng and Shang, Lifeng and others},
  journal={arXiv preprint arXiv:2401.17167},
  year={2024}
}

@article{wang2022text,
  title={Text embeddings by weakly-supervised contrastive pre-training},
  author={Wang, Liang and Yang, Nan and Huang, Xiaolong and Jiao, Binxing and Yang, Linjun and Jiang, Daxin and Majumder, Rangan and Wei, Furu},
  journal={arXiv preprint arXiv:2212.03533},
  year={2022}
}

@article{schick2023toolformer,
  title={Toolformer: Language models can teach themselves to use tools},
  author={Schick, Timo and Dwivedi-Yu, Jane and Dess{\`\i}, Roberto and Raileanu, Roberta and Lomeli, Maria and Hambro, Eric and Zettlemoyer, Luke and Cancedda, Nicola and Scialom, Thomas},
  journal={Advances in Neural Information Processing Systems},
  volume={36},
  pages={68539--68551},
  year={2023}
}

@article{yang2025qwen3,
  title={Qwen3 technical report},
  author={Yang, An and Li, Anfeng and Yang, Baosong and Zhang, Beichen and Hui, Binyuan and Zheng, Bo and Yu, Bowen and Gao, Chang and Huang, Chengen and Lv, Chenxu and others},
  journal={arXiv preprint arXiv:2505.09388},
  year={2025}
}

@article{achiam2023gpt,
  title={Gpt-4 technical report},
  author={Achiam, Josh and Adler, Steven and Agarwal, Sandhini and Ahmad, Lama and Akkaya, Ilge and Aleman, Florencia Leoni and Almeida, Diogo and Altenschmidt, Janko and Altman, Sam and Anadkat, Shyamal and others},
  journal={arXiv preprint arXiv:2303.08774},
  year={2023}
}

@article{wang2023plan,
  title={Plan-and-solve prompting: Improving zero-shot chain-of-thought reasoning by large language models},
  author={Wang, Lei and Xu, Wanyu and Lan, Yihuai and Hu, Zhiqiang and Lan, Yunshi and Lee, Roy Ka-Wei and Lim, Ee-Peng},
  journal={arXiv preprint arXiv:2305.04091},
  year={2023}
}

@article{hariharan2025plan,
  title={Plan Verification for LLM-Based Embodied Task Completion Agents},
  author={Hariharan, Ananth and Dongre, Vardhan and Hakkani-T{\"u}r, Dilek and Tur, Gokhan},
  journal={arXiv preprint arXiv:2509.02761},
  year={2025}
}

@article{wu2023autogen,
  title={Autogen: Enabling next-gen llm applications via multi-agent conversation framework},
  author={Wu, Qingyun and Bansal, Gagan and Zhang, Jieyu and Wu, Yiran and Zhang, Shaokun and Zhu, Erkang and Li, Beibin and Jiang, Li and Zhang, Xiaoyun and Wang, Chi},
  journal={arXiv preprint arXiv:2308.08155},
  volume={3},
  number={4},
  year={2023}
}

@article{brown2020language,
  title={Language models are few-shot learners},
  author={Brown, Tom and Mann, Benjamin and Ryder, Nick and Subbiah, Melanie and Kaplan, Jared D and Dhariwal, Prafulla and Neelakantan, Arvind and Shyam, Pranav and Sastry, Girish and Askell, Amanda and others},
  journal={Advances in neural information processing systems},
  volume={33},
  pages={1877--1901},
  year={2020}
}

@article{ahn2022can,
  title={Do as i can, not as i say: Grounding language in robotic affordances},
  author={Ahn, Michael and Brohan, Anthony and Brown, Noah and Chebotar, Yevgen and Cortes, Omar and David, Byron and Finn, Chelsea and Fu, Chuyuan and Gopalakrishnan, Keerthana and Hausman, Karol and others},
  journal={arXiv preprint arXiv:2204.01691},
  year={2022}
}

@article{valmeekam2023planning,
  title={On the planning abilities of large language models-a critical investigation},
  author={Valmeekam, Karthik and Marquez, Matthew and Sreedharan, Sarath and Kambhampati, Subbarao},
  journal={Advances in Neural Information Processing Systems},
  volume={36},
  pages={75993--76005},
  year={2023}
}

@article{shinn2023reflexion,
  title={Reflexion: Language agents with verbal reinforcement learning, 2023},
  author={Shinn, Noah and Cassano, Federico and Labash, Beck and Gopinath, Ashwin and Narasimhan, Karthik and Yao, Shunyu},
  journal={URL https://arxiv. org/abs/2303.11366},
  volume={1},
  year={2023}
}

@article{liu2024toolnet,
  title={Toolnet: Connecting large language models with massive tools via tool graph},
  author={Liu, Xukun and Peng, Zhiyuan and Yi, Xiaoyuan and Xie, Xing and Xiang, Lirong and Liu, Yuchen and Xu, Dongkuan},
  journal={arXiv preprint arXiv:2403.00839},
  year={2024}
}

@article{wu2024toolplanner,
  title={ToolPlanner: A tool augmented LLM for multi granularity instructions with path planning and feedback},
  author={Wu, Qinzhuo and Liu, Wei and Luan, Jian and Wang, Bin},
  journal={arXiv preprint arXiv:2409.14826},
  year={2024}
}

@article{ji2023survey,
  title={Survey of hallucination in natural language generation},
  author={Ji, Ziwei and Lee, Nayeon and Frieske, Rita and Yu, Tiezheng and Su, Dan and Xu, Yan and Ishii, Etsuko and Bang, Ye Jin and Madotto, Andrea and Fung, Pascale},
  journal={ACM computing surveys},
  volume={55},
  number={12},
  pages={1--38},
  year={2023},
  publisher={ACM New York, NY}
}

@article{lumer2025graph,
  title={Graph RAG-Tool Fusion},
  author={Lumer, Elias and Basavaraju, Pradeep Honaganahalli and Mason, Myles and Burke, James A and Subbiah, Vamse Kumar},
  journal={arXiv preprint arXiv:2502.07223},
  year={2025}
}

@article{kambhampati2024llms,
  title={Llms can't plan, but can help planning in llm-modulo frameworks},
  author={Kambhampati, Subbarao and Valmeekam, Karthik and Guan, Lin and Verma, Mudit and Stechly, Kaya and Bhambri, Siddhant and Saldyt, Lucas and Murthy, Anil},
  journal={arXiv preprint arXiv:2402.01817},
  year={2024}
}

@article{gu2024survey,
  title={A survey on llm-as-a-judge},
  author={Gu, Jiawei and Jiang, Xuhui and Shi, Zhichao and Tan, Hexiang and Zhai, Xuehao and Xu, Chengjin and Li, Wei and Shen, Yinghan and Ma, Shengjie and Liu, Honghao and others},
  journal={The Innovation},
  year={2024},
  publisher={Elsevier}
}

@article{wang2024officebench,
  title={Officebench: Benchmarking language agents across multiple applications for office automation},
  author={Wang, Zilong and Cui, Yuedong and Zhong, Li and Zhang, Zimin and Yin, Da and Lin, Bill Yuchen and Shang, Jingbo},
  journal={arXiv preprint arXiv:2407.19056},
  year={2024}
}

@article{wu2023visual,
  title={Visual chatgpt: Talking, drawing and editing with visual foundation models},
  author={Wu, Chenfei and Yin, Shengming and Qi, Weizhen and Wang, Xiaodong and Tang, Zecheng and Duan, Nan},
  journal={arXiv preprint arXiv:2303.04671},
  year={2023}
}

@article{yang2023mm,
  title={Mm-react: Prompting chatgpt for multimodal reasoning and action},
  author={Yang, Zhengyuan and Li, Linjie and Wang, Jianfeng and Lin, Kevin and Azarnasab, Ehsan and Ahmed, Faisal and Liu, Zicheng and Liu, Ce and Zeng, Michael and Wang, Lijuan},
  journal={arXiv preprint arXiv:2303.11381},
  year={2023}
}

@article{lewis2020retrieval,
  title={Retrieval-augmented generation for knowledge-intensive nlp tasks},
  author={Lewis, Patrick and Perez, Ethan and Piktus, Aleksandra and Petroni, Fabio and Karpukhin, Vladimir and Goyal, Naman and K{\"u}ttler, Heinrich and Lewis, Mike and Yih, Wen-tau and Rockt{\"a}schel, Tim and others},
  journal={Advances in neural information processing systems},
  volume={33},
  pages={9459--9474},
  year={2020}
}

\appendix

\section{Details of Perturbation-based Supervision}
\label{app:perturbation}

\subsection{Perturbation Operators}
\label{app:perturbation_operators}

Given a request $r$ and its ground truth plan graph $G^{\mathrm{gt}}_r$, we generate a set of type-executable but semantically corrupted graphs $\{G^{\mathrm{pert}}_{r}\}$ together with an operation log $\mathrm{ops}(G^{\mathrm{pert}}_{r})$ for each sample, which is then used to derive graph/node/edge supervision targets.

\subsubsection{Sampling numbers and operation budgets.}
For each $G^{\mathrm{gt}}_r$, we sample the number of perturbed graphs as
\begin{equation}
C \sim \mathrm{Categorical}(\{2,3,4\};\ [0.25,0.50,0.25]).
\end{equation}

For each perturbed graph, we sample an operation budget:
\begin{equation}
B \sim \mathrm{Categorical}(\{1,2,3\};\ [0.60,0.30,0.10]).
\end{equation}
This distribution is skewed toward small $B$ because realistic plans often fail due to one primary issue, while still allowing occasional multi-error cases to improve robustness. For each operation, we sample the operator family with equal probability:
$\Pr(\textsc{Wrong Tool}) \\
=\Pr(\textsc{Missing Step})=0.50$.

\subsubsection{Wrong Tool.}
Given a ground truth plan graph $G^{\mathrm{gt}}_r=(V_r,E_r)$, we randomly sample a node $v_i=(t_i,s_i)\in V_r$ and \textsc{REPLACE} its tool $t_i$ with an alternative tool $t_i'$, while keeping the step text $s_i$ unchanged.
To form hard negatives, $t_i'$ is sampled preferentially from the similar tool neighborhood $\mathcal{N}_K(t_i)$.
To avoid trivially infeasible negatives, we require the replacement to preserve interface-level connectivity under the dependency graph, i.e., it must be connectable to both adjacent tools:
\begin{equation}
t_i' \in 
\mathcal{N}^{\text{tool}}_{\mathrm{out}}(t_{i-1})
\ \cap\
\mathcal{N}^{\text{tool}}_{\mathrm{in}}(t_{i+1}),
\label{eq:wt-io-filter}
\end{equation}
where endpoint nodes only check the existing side.

We then draw $t_i'$ from two complementary subsets to control perturbation hardness:
(i) Semantic-neighbor confusion. With probability $0.75$, we sample $t_i'$ from $\mathcal{N}_K(t_i)$ after applying the connectivity filter in Eq.~\eqref{eq:wt-io-filter}, producing hard negatives that are type-executable and semantically close to $t_i$, which resembles fine-grained confusions frequently made by real planners.
(ii) Mild noise. With probability $0.25$, we sample $t_i'$ from the remaining feasible tools that pass Eq.~\eqref{eq:wt-io-filter} but are not in $\mathcal{N}_K(t_i)$.
This injects lightweight noise while still respecting the executability constraint, covering less frequent but possible tool selection mistakes.
If the feasible set is empty, we resample $i$ until a valid replacement is found.

\subsubsection{Missing Step.}
When \textsc{Missing Step} is chosen, we sample a consecutive directed subchain
$t_u\rightarrow t_{u+1}\rightarrow\cdots\rightarrow t_v$ in $G^{\mathrm{gt}}_r$ while keeping the perturbed plan type-executable under the dependency graph. Then we sample the subtype with
$\Pr(\textsc{DROP})=\Pr(\textsc{COMPRESS})=0.50$.

\textbf{DROP(span).}
We sample the span length:
\begin{equation}
m \sim \mathrm{Categorical}(\{1,2,3,4,\ge 5\};\ [0.55,0.25,0.10,0.07,0.03]),
\end{equation}
where the distribution is skewed toward short spans to reflect that real planners more often skip one or two critical intermediate steps, while still assigning a small probability to longer spans to generate harder cases with more severe missing step errors.
Then delete $m$ consecutive nodes and directly connect the endpoint tools by a shortcut edge $t_u\rightarrow t_v$, (or $\textsc{Start}\rightarrow t_v$ when the span starts from the beginning).
When $t_u\neq \textsc{Start}$, we require the shortcut edge to be type-executable, i.e., $t_v \in \mathcal{N}^{\text{tool}}_{\mathrm{out}}(t_u)$; otherwise we resample the span (or $m$) until the shortcut is feasible. To make the perturbation realistic, we mix two sampling modes over feasible spans. With probability $0.75$, we sample spans proportionally to a motif weight derived from tool sequence counts $f_n(\cdot)$ (to favor locally plausible shortcuts that a planner might produce), and with probability $0.25$ we sample uniformly among feasible spans to cover long-tail omissions.

\textbf{COMPRESS(span$\rightarrow$1).}
We select a consecutive span and replace it by a single shortcut tool $t^*$, forming $t_u\rightarrow t^*\rightarrow t_v$ (or $\textsc{Start}\rightarrow t^*\rightarrow t_v$ when the span starts from \textsc{Start}).

In implementation, we restrict the span length to $m\in\{2,3,4\}$ and sample a feasible span uniformly from all candidate spans of these lengths.
The shortcut tool $t^*$ is enumerated from the dependency graph as a bridge candidate that preserves type executability:
\begin{equation}
t^* \in \mathcal{N}^{\mathrm{tool}}_{\mathrm{out}}(t_u)\cap \mathcal{N}^{\mathrm{tool}}_{\mathrm{in}}(t_v),
\end{equation}
and for the spans starts from \textsc{Start} we drop the left constraint and only require $t^*\in \mathcal{N}^{\text{tool}}_{\mathrm{in}}(t_v)$.

To prefer plausible but improper compressions, we rank candidates using the span text
$s_{\mathrm{span}}$ (concatenation of the deleted step texts) and sample:
\begin{equation}
\Pr(t^*) \propto \exp\!\big(g(s_{\mathrm{span}},t^*)\big),
\end{equation}
so that $t^*$ tends to be semantically aligned with the span while still representing an over-generalized merge.
If no feasible $t^*$ exists, we resample the span.

The two perturbation types can be applied independently or composed multiple times on the same $G^{\mathrm{gt}}_r$, yielding candidate graphs with varying severities and error locations.

\subsection{Details of Supervision Signals}
\label{app:cost}

\textbf{Graph-level target.}
For each $G^{\mathrm{pert}}_r$, we compute a non-negative perturbation cost
$c(G^{\mathrm{pert}}_r) = \allowbreak\sum_{o\in\mathrm{ops}(G^{\mathrm{pert}}_r)}\eta(o)$,
where $\eta(o)$ is the cost of executing one perturbation operation $o$. We then map it to the graph-level target $y(G^{\mathrm{pert}}_r)=\exp\!\big(-c(G^{\mathrm{pert}}_r)/\tau\big)$ as in Eq.~\ref{eq:soft-target}.
In implementation, we use the following operation-type costs:

\textbf{REPLACE.}
For a replacement at node $v_i=(t_i,s_i)$ where the tool becomes $t_i'$, we set
\begin{equation}
\eta(\textsc{REPLACE})=\big(1- g(s_i,t_i')\big).
\end{equation}
This assigns smaller cost to harder near-synonym confusions that still align well with the step text.

\textbf{DROP(span).}
Let $D$ be the deleted tool set and $m=|D|$. We use
\begin{equation}
\eta(\textsc{DROP})=m\cdot \sum_{t\in D}\max\big(0,s(r,t)\big),
\end{equation}
where $s(r,t)$ is the semantic similarity between the request and the tool representation.
This penalizes dropping longer spans and dropping tools that are relevant to the request.

\textbf{COMPRESS(span$\rightarrow$1).}
For a span of length $m$ compressed into a shortcut tool $t^*$ with span text $s_{\mathrm{span}}$
(concatenation of the deleted step texts), we set
\begin{equation}
\eta(\textsc{COMPRESS})=(m-1)\cdot \big(1-g(s_{\mathrm{span}},t^*)).
\end{equation}
This assigns larger cost to longer compressions and to shortcuts that poorly cover the original span.

\subsection{Details of Two-stage Training}
\label{app:two_stage_training}

For each request $r$, we construct a training set that contains the ground truth plan graph $G^{\mathrm{gt}}_r$ and a set of perturbed graphs $\{G^{\mathrm{pert}}_{r}\}$.
Each graph $G_r$ is associated with (i) a graph-level target $y(G)\in(0,1]$, (ii) node-level target $\{\ell^{\mathrm{node}}_v\}_{v\in V_r}$, and (iii) edge-level target $\{\ell^{\mathrm{edge}}_{uv}\}_{(u,v)\in E_r}$.

We train the verifier in two stages: Stage~I learns stable graph-level scoring, and Stage~II learns fine-grained node/edge diagnosis with minimal drift of the global capability.

\textbf{Stage I: Graph-level training.}
We supervise the graph-level logit $z(G_r)$ with the graph-level soft target $y(G_r)$ using a binary cross-entropy regression loss:
\begin{equation}
\label{eq:loss-graph}
\mathcal{L}_{\mathrm{graph}}=\mathrm{BCEWithLogits}\big(z(G_r),\ y(G_r)\big).
\end{equation}
In addition, for the same request $r$ we typically have multiple graphs with different perturbation costs $c(\cdot)$, which naturally induce an ordering: $\mathcal{P}(r)=\{(i,j)\mid c(G_{r,i})<c(G_{r,j})\}$ 
and apply a margin ranking loss:
\begin{equation}
\label{eq:app_loss_rank}
\begin{aligned}
\mathcal{L}_{\mathrm{rank}} &=
\sum_{(i,j)\in\mathcal{P}(r)}
\max\!\Big(0,\ m_{ij}-z(G_{r,i})+z(G_{r,j})\Big),
\end{aligned}
\end{equation}
where $m_{ij}=0.2(c(G_{r,j})-c(G_{r,i}))$, this formulation encourages the verifier to assign higher scores to less corrupted graphs, while enforcing larger separation for larger corruption gaps.
The overall Stage~I objective is
\begin{equation}
\label{eq:app_loss_stage1}
\mathcal{L}_{\mathrm{stage1}}
=
\mathcal{L}_{\mathrm{rank}}
+
\lambda_{\mathrm{graph}}\mathcal{L}_{\mathrm{graph}},
\end{equation}
where $\lambda_{\mathrm{graph}}$ balances calibration against ranking consistency.

\textbf{Stage II: Node/Edge-level training.}
Stage~II focuses on local diagnosis signals.
We freeze the most encoder parameters learned in Stage~I, and fine-tune only the last GNN layer together with the node-/edge-level prediction heads.
For each node $v\in V_r$ and each directed edge $(u,v)\in E_r$, we compute logits $z_r^V(v)$ and $z_r^E(u,v)$, and optimize them with weighted binary cross-entropy losses:
\begin{equation}
\begin{aligned}
\mathcal{L}_{\mathrm{node}}
&=
\sum_{v\in V_r}\mathrm{WBCEWithLogits}\big(z_r^V(v),\ \ell^{\mathrm{node}}_v;\ \omega_{\mathrm{node}}\big),\\
\label{eq:loss-local}
\mathcal{L}_{\mathrm{edge}}
&=
\sum_{(u,v)\in E_r}
\mathrm{WBCEWithLogits}\big(z_r^E(u,v),\ \ell^{\mathrm{edge}}_{uv};\ \omega_{\mathrm{edge}}\big),
\end{aligned}
\end{equation}
since positive targets are rarer than negative ones, we set $\omega_{\mathrm{node}}$ and $\omega_{\mathrm{edge}}>1$ to mitigate class imbalance. Both weights are computed automatically from the positive/negative target counts in the training data.
The overall Stage~II objective is
\begin{equation}
\label{eq:loss-stage2}
\mathcal{L}_{\mathrm{stage2}}=\mathcal{L}_{\mathrm{node}}+\lambda_{\mathrm{edge}}\mathcal{L}_{\mathrm{edge}},
\end{equation}
where $\lambda_{\mathrm{edge}}$ controls the strength of edge-level diagnosis relative to node-level diagnosis.

\newpage

\section{Details of Local Correction}
\label{app:llm_correction}
This appendix details the implementation of our verification-guided local correction module.
Given the verifier outputs on a plan graph, we expose a small set of high-risk nodes/edges together with constrained candidate tools, and require the LLM to return minimal, schema-valid edit operations.

\subsection{Candidate Tools}
\label{app:candidate_tools}
In this section, we detail how we construct tool candidate sets from $G_{\text{tool}}$ for node replacement and edge insertion during local correction, and how we rank and select the final candidates.

\textbf{Replacement candidates (node-level).}
For any high-risk node $v_i=(t_i,s_i)\in\mathcal{V}_{\mathrm{edit}}$, we select candidates from the dependency graph $G_{\text{tool}}$ and the similar tool neighborhood:
\begin{equation}
\mathcal{C}^{\mathrm{rep}}(v_i)
=
\mathcal{N}_K(t_i)
\cap
\mathcal{N}^{\text{tool}}_{\mathrm{out}}(t_{i-1})
\cap
\mathcal{N}^{\text{tool}}_{\mathrm{in}}(t_{i+1}).
\end{equation}
When $v_i$ is a root node (zero in-degree), we treat $t_{i-1}$ as \textsc{Start} and only keep
$\mathcal{N}^{\text{tool}}_{\mathrm{in}}(t_{i+1})$ in the above filtering.
We rank $t\in\mathcal{C}^{\mathrm{rep}}(v_i)$ by
\begin{equation}
\label{eq:score-rep}
\mathrm{score}_{\mathrm{rep}}(t)=g(s_i,t) + s(r,t).
\end{equation}
Finally return the Top-3 candidates.

\textbf{Insertion candidates (edge-level).}
For any high-risk edge $(u,v)\in\mathcal{E}_{\mathrm{edit}}$, we enumerate bridging tools from the dependency graph $G_{\text{tool}}$:
\begin{equation}
\label{eq:cand-ins}
\mathcal{C}^{\mathrm{ins}}(u,v)
=  
\mathcal{N}^{\text{tool}}_{\mathrm{out}}(t_u)
\cap
\mathcal{N}^{\text{tool}}_{\mathrm{in}}(t_v).
\end{equation}
When $u=\textsc{Start}$, we drop the constraint induced by $\mathcal{N}^{\text{tool}}_{\mathrm{out}}(t_u)$ and only require
$t\in\mathcal{N}^{\text{tool}}_{\mathrm{in}}(t_v)$.
We rank $t\in\mathcal{C}^{\mathrm{ins}}(u,v)$ by
\begin{equation}
\label{eq:score-ins}
\mathrm{score}_{\mathrm{ins}}(t)=0.8s(r,t) + 0.2\left(\log\!\big(1+f_2(t_u,t)\big)+\log\!\big(1+f_2(t,t_v)\big)\right),
\end{equation}
where $f_2(\cdot,\cdot)$ is the frequency of length-2 tool sequences.
Finally return the Top-3 candidates.
We choose 0.8 to prioritize request--tool relevance when inserting bridging tools (to avoid introducing irrelevant steps), while the co-occurrence statistics $f_2(\cdot,\cdot)$ serve as an auxiliary signal to favor more common local transitions.

\subsection{Threshold Selection}
\label{app:threshold_selection}
The thresholds $\tau_G,\tau_V,\tau_E$ directly control whether correction is triggered and how large the editable region is.
We calibrate them on the validation set in two steps: (i) Sweep $\tau_V$ and $\tau_E$ using $P^V_r(v)$ and $P^E_r(u,v)$  over a grid and evaluate them as binary detectors of ground truth error locations:
a node (or edge) is predicted as positive if its risk exceeds the threshold.
We then choose the threshold that maximizes the validation F1 score; (ii) With $\tau_V$ and $\tau_E$ fixed, we tune the graph-level acceptance threshold $\tau_G$ for deciding whether to invoke the LLM corrector.
We sweep $\tau_G$ on the validation set and apply local correction only to samples with $S_r<\tau_G$.
We select $\tau_G$ that maximizes the end-to-end validation accuracy of the final plan,
reflecting the trade-off between correcting low quality plans and avoiding over editing reasonably good plans.

\begin{figure*}[t]
\centering
\begin{minipage}{0.78\textwidth}
\lstset{style=promptbox}
\begin{lstlisting}
# DEPENDENCY GRAPH #
{{ dependency_graph }}

# USER REQUEST #
{{ user_request }}

# CURRENT PLAN (JSON) #
{ "task_steps": {{ task_steps }}, "task_nodes": {{ task_nodes }}, "task_links": {{ task_links }}}

# GNN EVALUATION #
The GNN is a plan evaluator and reports:
- Graph score $S \in (0,1)$: higher suggests stronger alignment with the user request.
- Node risk $ \in (0,1)$: higher indicates a node may use an incorrect tool.
- Edge risk $ \in (0,1)$: higher indicates a edge (including START $\rightarrow$ root node and tool-to-tool edges) may be incomplete and need an inserted tool.

# TOP RISK NODES #
{node_diag_str}

# TOP RISK EDGES #
{edge_diag_str}

# CANDIDATES (node_id -> candidate_id -> tool) #
{{ node_candidates_json }}

# CANDIDATES (edge_id -> candidate_id -> tool) #
{{ edge_candidates_json }}

# GOAL #
You are a plan refinement assistant. Analyze the user request, the current plan, and the GNN diagnostics to decide whether and how to improve the plan so it better satisfies the request.

# COMMON FAILURE MODES #
1) Wrong Tool: a tool is semantically similar but incorrect $\rightarrow$ replace the node.
2) Missing Step: especially missing preprocessing $\rightarrow$ insert a tool on the risky edge.

# TASK #
Based on the user request, the current plan, and the GNN diagnostics, first decide whether any change is necessary. If no change is needed, return empty edits. If changes are needed, select replacement/insertion tools only from the provided candidates and propose minimal edits. If improvements are not clear with the provided candidates, do not modify. Return EDIT OPERATIONS only (no analysis text). It is valid to return no edits.

# RULES #
1. Output JSON ONLY (no extra text).
2. Modify at most 3 places and do not use the same candidate tool in multiple edits.
3. Allowed ops:
   - replace_on_node(node_id, candidate_id, step)
   - insert_on_edge(edge_id, candidate_id, step)
   - no_change()
4. candidate_id must be an integer index from the candidate list. 
5. Each node/edge may remain unchanged; prefer fewer edits and only change when necessary.
6. For every edit, provide a new step text aligned with the chosen tool and request and keep steps aligned 1-to-1 with nodes (same count, same order).
7. The updated steps/tools should solve the request better than the current plan; otherwise do not modify.

# OUTPUT FORMAT (minimal edits) #
{{
  "edits": [
    {{"op":"replace_on_node","node_id":0,"candidate_id":1,"step":"Step 1: ..."}},
    {{"op":"insert_on_edge","edge_id":0,"candidate_id":2,"step":"Step 2: ..."}}
  ]
}}
If the current workflow is already optimal, return: {{"edits":[]}}

# RESULT #

\end{lstlisting}
\end{minipage}
\vspace{-0.3em}
\caption{Prompt template for verification-guided local correction.}
\label{fig:prompt_local_correction}
\vspace{-0.8em}
\end{figure*}

\begin{figure*}[t]
\centering
\begin{minipage}{0.78\textwidth}
\lstset{style=promptbox}
\begin{lstlisting}
# USER REQUEST #
{{ user_request }}

# ERROR #
{{ error_msg }}

# CANDIDATES (node_id -> candidate_id -> tool) #
{{ node_candidates_json }}

# CANDIDATES (edge_id -> candidate_id -> tool) #
{{ edge_candidates_json }}

# TASK #
Fix the edit operations to satisfy all constraints.
Only output:
{{"edits":[
  {{"op":"replace_on_node","node_id":0,"candidate_id":1,"step":"Step 1: ..."}},
  {{"op":"insert_on_edge","edge_id":0,"candidate_id":2,"step":"Step 2: ..."}}
]}}
Or {{\"edits\":[]}} / no_change().

# RESULT #

\end{lstlisting}
\end{minipage}
\vspace{-0.3em}
\caption{Retry prompt template for schema-valid edit operations.}
\label{fig:prompt_local_correction_retry}
\vspace{-0.8em}
\end{figure*}

\subsection{Inputs and Outputs}
\label{app:llm_inputs_outputs}
\textbf{Inputs.}
Given request $r$ and the current plan graph $G_r=(V_r,E_r)$, the verifier outputs a graph-level score $S_r$, node-level risk
$\mathbf{p}^V_r$, and edge-level  risk $\mathbf{p}^E_r$.
Editable regions $\mathcal{V}_{\text{edit}}$ and $\mathcal{E}_{\text{edit}}$ are obtained by thresholding risks as in Eq.~\eqref{eq:editable-region}.
When local correction is triggered (i.e., $S_r<\tau_G$), we rank nodes in $\mathcal{V}_{\text{edit}}$ by $P^V_r(v)$ and edges in $\mathcal{E}_{\text{edit}}$ by $P^E_r(u,v)$, and expose only Top-3 risky locations to the LLM.

For each exposed high-risk node $v_i\in\mathcal{V}_{\text{edit}}$, we provide a Top-3 replacement list
$\mathcal{C}^{\mathrm{rep}}(v_i)$; for each exposed high-risk edge $(u,v)\in\mathcal{E}_{\text{edit}}$, we provide a Top-$3$ insertion list
$\mathcal{C}^{\mathrm{ins}}(u,v)$.

\textbf{Outputs.}
The LLM is required to output \emph{JSON only} with a single field \texttt{edits}.
Each element in \texttt{edits} specifies one edit operation:
\texttt{replace\_on\_node} or \texttt{insert\_on\_edge}.
Returning an empty list (\texttt{"edits": []}) indicates no change.

We use the following minimal schema:
\begin{verbatim}
{
  "edits": [
    {"op":"replace_on_node", "node_id": i, 
          "candidate_id": j, "step": "Step k: ..."},
    {"op":"insert_on_edge", "edge_id":  i, 
          "candidate_id": j, "step": "Step k: ..."}
  ]
}
\end{verbatim}
where \texttt{node\_id} indexes a tool node in the current plan graph, and \texttt{edge\_id} indexes an exposed risky edge (including the virtual \textsc{Start}$\rightarrow$root edge when applicable).
\texttt{candidate\_id} must be an integer index into the corresponding.

Before applying edits, we enforce the following constraints:
(1) Edit budget: at most 3 edits;
(2) Index validity: \texttt{node}/\texttt{edge\_id} must refer to an exposed location;
(3) Candidate validity: \texttt{candidate\_id} must be within the Top-3 candidate list of that location;
(4) No repeated tool: the same candidate tool cannot be used in multiple edits;
(5) Step text: each edit must include a non-empty step description aligned with the chosen tool.
If any constraint is violated or the output cannot be parsed as JSON, we trigger a single retry.

We provide the following prompt template to drive the LLM to perform constrained local edits based on the verifier diagnostics and the candidate tools from the dependency graph (Figure~\ref{fig:prompt_local_correction}).

\subsection{Retry and Acceptance Rules}
\label{app:llm_retry}
If the LLM output cannot be parsed as valid JSON or violates any constraint in Sec.~\ref{app:llm_inputs_outputs},
we prompt the LLM once again with (i) the request $r$, (ii) an error message summarizing the violated constraint(s),
and (iii) the same exposed candidate lists, asking it to \emph{only} fix the edit operations and return JSON.

After applying the edits to obtain a corrected plan graph $G'_r$, we re-run the verifier to compute $S'_r$.
We accept the correction only if it improves the graph-leve score, i.e., $S'_r>S_r$; otherwise we keep the original plan.
This conservative acceptance rule prevents the corrector from making unnecessary or harmful changes when the constrained edits do not yield a clearer improvement.

The retry uses a lightweight fix-up prompt that only exposes the error message and the same candidate lists, as shown in Figure~\ref{fig:prompt_local_correction_retry}.

\section{Dataset Details}
\label{app:datasets}

\textbf{TaskBench}~\cite{shen2024taskbench} is a multi-domain benchmark for tool-based task planning.
For each domain, it defines a dependency graph, where nodes are tools/APIs and directed edges represent dependencies between tools.
Each instance pairs a natural language user request with decomposed step texts and a ground truth tool invocation graph (as nodes and dependency links), generated under controlled graph structure templates (e.g., single-node, chain, and directed acyclic graph) and filtered by quality checks. 
In our experiments, we use the three TaskBench domains (HuggingFace, Multimedia, and DailyLife), and directly treat each instance as a plan graph consistent with prior graph-based planning work. We follow the official formatting and preprocessing conventions.

\textbf{UltraTool}~\cite{huang2024planning} targets large-scale tool use planning with broad tool coverage and multi-step tool invocation graphs. 
Each instance contains a user request and a ground truth tool invocation plan. 
Following the same preprocessing protocol of GNN4Plan~\cite{wu2024can}, we derive a challenging benchmark with a larger global dependency graph. 
Tool descriptions are refined under the same protocol to ensure semantic consistency.

\twocolumn[{
\centering
\captionof{table}{Performance comparison across four datasets on Qwen3: Node-F1, Link-F1, and Accuracy are reported in \%. The best results are highlighted in boldface, and the second-best results are underlined.}
\label{tab:main_results2}
\setlength{\tabcolsep}{5pt}
\renewcommand{\arraystretch}{1.25}
\begin{adjustbox}{max width=\textwidth}
\begin{tabular}{p{7em}l ccc ccc ccc ccc}

\toprule
\multicolumn{1}{p{7em}}{\multirow{2}{*}{}} & \multirow{2}{*}{\textbf{Method}}
& \multicolumn{3}{c}{\textbf{HuggingFace}}
& \multicolumn{3}{c}{\textbf{Multimedia}}
& \multicolumn{3}{c}{\textbf{DailyLife}}
& \multicolumn{3}{c}{\textbf{UltraTool}} \\
\cmidrule(lr){3-5} \cmidrule(lr){6-8} \cmidrule(lr){9-11} \cmidrule(lr){12-14}
& & \textit{n-F1}$\uparrow$ & \textit{l-F1}$\uparrow$ & \textit{Acc}$\uparrow$
  & \textit{n-F1}$\uparrow$ & \textit{l-F1}$\uparrow$ & \textit{Acc}$\uparrow$
  & \textit{n-F1}$\uparrow$ & \textit{l-F1}$\uparrow$ & \textit{Acc}$\uparrow$
  & \textit{n-F1}$\uparrow$ & \textit{l-F1}$\uparrow$ & \textit{Acc}$\uparrow$ \\
\midrule

\multicolumn{1}{c}{\multirow{5}{*}{\textbf{Direct}}}
& \cellcolor{gray!15} Raw & \cellcolor{gray!15} 82.41 & \cellcolor{gray!15} 58.63 & \cellcolor{gray!15} 34.20 & \cellcolor{gray!15} 89.40 & \cellcolor{gray!15} 71.56 & \cellcolor{gray!15} 53.40 & \cellcolor{gray!15} \textbf{97.59} & \cellcolor{gray!15} \underline{67.04} & \cellcolor{gray!15} \underline{54.20} & \cellcolor{gray!15} 64.84 & \cellcolor{gray!15} 37.67 & \cellcolor{gray!15} 31.20 \\
& \hspace*{2pt}+Refine & \hspace*{2pt}80.99 & \hspace*{2pt}56.53 & \hspace*{2pt}33.20 & \hspace*{2pt}89.52 & \hspace*{2pt}72.29 & \hspace*{2pt}54.40 & \hspace*{2pt}96.93 & \hspace*{2pt}62.56 & \hspace*{2pt}49.20 & \hspace*{2pt}\textbf{73.31} & \hspace*{2pt}30.24 & \hspace*{2pt}23.40 \\
& \cellcolor{gray!15} +VeriCoder & \cellcolor{gray!15} 81.47 & \cellcolor{gray!15} 57.56 & \cellcolor{gray!15} 34.20 & \cellcolor{gray!15} 89.33 & \cellcolor{gray!15} 72.50 & \cellcolor{gray!15} 53.80 & \cellcolor{gray!15} 97.05 & \cellcolor{gray!15} 44.49 & \cellcolor{gray!15} 29.46 & \cellcolor{gray!15} 72.34 & \cellcolor{gray!15} 29.77 & \cellcolor{gray!15} 23.05 \\
& \hspace*{2pt}+VeriPlan & \hspace*{2pt}\underline{82.47} & \hspace*{2pt}\underline{58.92} & \hspace*{2pt}\underline{34.60} & \hspace*{2pt}\underline{89.57} & \hspace*{2pt}\underline{72.65} & \hspace*{2pt}\underline{54.80} & \hspace*{2pt}\textbf{97.59} & \hspace*{2pt}\underline{67.04} & \hspace*{2pt}\underline{54.20} & \hspace*{2pt}71.91 & \hspace*{2pt}\underline{44.13} & \hspace*{2pt}\underline{35.20} \\
& \cellcolor{gray!15} \textbf{+GNNVerifier} & \cellcolor{gray!15} \textbf{85.10} & \cellcolor{gray!15} \textbf{64.67} & \cellcolor{gray!15} \textbf{45.80} & \cellcolor{gray!15} \textbf{90.84} & \cellcolor{gray!15} \textbf{75.88} & \cellcolor{gray!15} \textbf{63.20} & \cellcolor{gray!15} \textbf{97.59} & \cellcolor{gray!15} \textbf{84.74} & \cellcolor{gray!15} \textbf{75.55} & \cellcolor{gray!15} \underline{72.81} & \cellcolor{gray!15} \textbf{45.21} & \cellcolor{gray!15} \textbf{37.47} \\
\midrule

\multicolumn{1}{c}{\multirow{5}{*}{\textbf{ReAct}}}
& \cellcolor{gray!15} Raw & \cellcolor{gray!15} \underline{81.89} & \cellcolor{gray!15} 56.90 & \cellcolor{gray!15} \underline{34.60} & \cellcolor{gray!15} 90.06 & \cellcolor{gray!15} 56.11 & \cellcolor{gray!15} 42.20 & \cellcolor{gray!15} \textbf{97.07} & \cellcolor{gray!15} 48.15 & \cellcolor{gray!15} 32.60 & \cellcolor{gray!15} 72.97 & \cellcolor{gray!15} 31.15 & \cellcolor{gray!15} 24.20 \\
& \hspace*{2pt}+Refine & \hspace*{2pt}78.20 & \hspace*{2pt}50.75 & \hspace*{2pt}28.00 & \hspace*{2pt}88.50 & \hspace*{2pt}53.76 & \hspace*{2pt}38.00 & \hspace*{2pt}94.17 & \hspace*{2pt}\underline{53.62} & \hspace*{2pt}\underline{33.20} & \hspace*{2pt}73.11 & \hspace*{2pt}34.12 & \hspace*{2pt}27.20 \\
& \cellcolor{gray!15} +VeriCoder & \cellcolor{gray!15} 78.39 & \cellcolor{gray!15} 51.31 & \cellcolor{gray!15} 26.60 & \cellcolor{gray!15} \underline{90.53} & \cellcolor{gray!15} 72.96 & \cellcolor{gray!15} 53.40 & \cellcolor{gray!15} 96.30 & \cellcolor{gray!15} 31.79 & \cellcolor{gray!15} 15.40 & \cellcolor{gray!15} 72.80 & \cellcolor{gray!15} 26.06 & \cellcolor{gray!15} 19.60 \\
& \hspace*{2pt}+VeriPlan & \hspace*{2pt}81.85 & \hspace*{2pt}\underline{56.95} & \hspace*{2pt}\underline{34.60} & \hspace*{2pt}90.15 & \hspace*{2pt}\underline{73.33} & \hspace*{2pt}\underline{54.80} & \hspace*{2pt}\textbf{97.07} & \hspace*{2pt}48.15 & \hspace*{2pt}32.60 & \hspace*{2pt}\underline{75.83} & \hspace*{2pt}\underline{36.48} & \hspace*{2pt}\underline{27.40} \\
& \cellcolor{gray!15} \textbf{+GNNVerifier} & \cellcolor{gray!15} \textbf{83.59} & \cellcolor{gray!15} \textbf{62.82} & \cellcolor{gray!15} \textbf{45.00} & \cellcolor{gray!15} \textbf{91.56} & \cellcolor{gray!15} \textbf{76.02} & \cellcolor{gray!15} \textbf{63.80} & \cellcolor{gray!15} \textbf{97.07} & \cellcolor{gray!15} \textbf{86.57} & \cellcolor{gray!15} \textbf{77.35} & \cellcolor{gray!15} \textbf{75.99} & \cellcolor{gray!15} \textbf{50.47} & \cellcolor{gray!15} \textbf{42.60} \\
\midrule

\multicolumn{1}{c}{\multirow{5}{*}{\textbf{GNN4Plan}}}
& \cellcolor{gray!15} Raw & \cellcolor{gray!15} \underline{80.56} & \cellcolor{gray!15} \underline{61.76} & \cellcolor{gray!15} \underline{44.20} & \cellcolor{gray!15} 87.25 & \cellcolor{gray!15} \underline{73.25} & \cellcolor{gray!15} \textbf{62.00} & \cellcolor{gray!15} \textbf{97.54} & \cellcolor{gray!15} \textbf{84.63} & \cellcolor{gray!15} \textbf{75.40} & \cellcolor{gray!15} 64.97 & \cellcolor{gray!15} 40.71 & \cellcolor{gray!15} 34.40 \\
& \hspace*{2pt}+Refine & \hspace*{2pt}77.22 & \hspace*{2pt}49.20 & \hspace*{2pt}23.80 & \hspace*{2pt}88.50 & \hspace*{2pt}70.15 & \hspace*{2pt}51.40 & \hspace*{2pt}94.09 & \hspace*{2pt}63.99 & \hspace*{2pt}49.60 & \hspace*{2pt}\textbf{73.98} & \hspace*{2pt}\underline{44.42} & \hspace*{2pt}\underline{36.40} \\
& \cellcolor{gray!15} +VeriCoder & \cellcolor{gray!15} 80.33 & \cellcolor{gray!15} 54.94 & \cellcolor{gray!15} 31.80 & \cellcolor{gray!15} \textbf{88.87} & \cellcolor{gray!15} 72.46 & \cellcolor{gray!15} 53.80 & \cellcolor{gray!15} 96.57 & \cellcolor{gray!15} 44.60 & \cellcolor{gray!15} 28.60 & \cellcolor{gray!15} 72.04 & \cellcolor{gray!15} 29.72 & \cellcolor{gray!15} 22.60 \\
& \hspace*{2pt}+VeriPlan & \hspace*{2pt}\underline{80.56} & \hspace*{2pt}\underline{61.76} & \hspace*{2pt}\underline{44.20} & \hspace*{2pt}87.25 & \hspace*{2pt}\underline{73.25} & \hspace*{2pt}\textbf{62.00} & \hspace*{2pt}\textbf{97.54} & \hspace*{2pt}\textbf{84.63} & \hspace*{2pt}\textbf{75.40} & \hspace*{2pt}66.04 & \hspace*{2pt}41.39 & \hspace*{2pt}35.20 \\
& \cellcolor{gray!15} \textbf{+GNNVerifier} & \cellcolor{gray!15} \textbf{83.30} & \cellcolor{gray!15} \textbf{63.94} & \cellcolor{gray!15} \textbf{46.00} & \cellcolor{gray!15} \underline{88.55} & \cellcolor{gray!15} \textbf{74.13} & \cellcolor{gray!15} \textbf{62.00} & \cellcolor{gray!15} \textbf{97.54} & \cellcolor{gray!15} \textbf{84.63} & \cellcolor{gray!15} \textbf{75.40} & \cellcolor{gray!15} \underline{73.11} & \cellcolor{gray!15} \textbf{45.78} & \cellcolor{gray!15} \textbf{38.68} \\
\bottomrule

\end{tabular}
\end{adjustbox}
\vspace{0.8em}
}]

\section{Additional Experimental Results}
\label{app:experimental_results_additional}
\subsection{Main Results on Qwen3}
\label{app:main_results_qwen3}

Table~\ref{tab:main_results2} reports the results on Qwen3-235B-A22B-Instruct-2507. Compared to the best baseline VeriPlan, the average relative improvements are 2.90\% / 7.77\% / 20.63\% on HuggingFace, 1.49\% / 3.10\% / 10.14\% on Multimedia, 0 / 28.09\% / 40.75\% on DailyLife, and 3.80\% / 15.95\% / 21.42\% on UltraTool (n-F1/l-F1/Acc). Across planners, the corresponding improvements are 1.41\% / 11.44\% / 24.17\% for Direct, 0.96\% / 28.37\% / 53.11\% for ReAct, and 3.35\% / 2.85\% / 2.44\% for GNN4Plan.
Across GPT-4o and Qwen3, we observe consistent trends: our graph-based verifier improves n-F1 and l-F1 in most planner--dataset combinations and yields the largest gains on task accuracy, indicating good generalization across backbone LLMs. On DailyLife, both backbones show near-saturated n-F1, while l-F1 and Acc improve substantially, suggesting remaining errors are mainly structural/link-related.

\subsection{Additional Ablation Studies}
\label{app:abalation_study}
Table~\ref{tab:ablation_results_p2} reports ablation results on DailyLife and UltraTool. Overall, the trends are consistent with HuggingFace and Multimedia: ablating any key component (graph-structured modeling, two-stage training, advanced node/edge attributes, or multi-granularity feedback) leads to performance degradation, and the full model achieves the best overall results across planners and metrics. In particular, removing the GNN or any feedback signal generally hurts link-level quality and task accuracy, confirming that structure-aware reasoning and complementary diagnostic feedback are both important for stable correction. A notable difference is DailyLife, where n-F1 is already near-saturated across variants (many settings share the same n-F1), so improvements mainly manifest in l-F1 and Acc, indicating limited headroom on node identification while structural/dependency correction remains the primary bottleneck.

\subsection{Additional Error Analysis}
\label{app:error_breakdown_react_gnn4plan}
Figures~\ref{fig:case_react} and~\ref{fig:case_gnn4plan} illustrate the distribution of five error types for the ReAct and GNN4Plan planners, respectively, comparing the raw plans (\textit{Before}) with the plans after correction by our method (\textit{After}). This analysis aims to verify the generalizability and effectiveness of our correction method across different planner frameworks.
    
\textbf{ReAct.}
Across most datasets, our method reduces planning errors, with clear improvements on \textit{Missing Tool} and \textit{Edge Fail}, suggesting that verification-guided local correction can effectively insert missing intermediate steps and fix semantically incorrect yet type-compatible transitions.
Notably, on DailyLife, the \textit{Wrong Tool} count remains unchanged and is already small, indicating that its errors are dominated by other types and thus leave limited room for tool-replacement gains; \textit{Dependency Error} also stays negligible with a minor fluctuation (2$\rightarrow$3), likely due to its small error mass.

\textbf{GNN4Plan.}
A notable difference is that \textit{Dependency Error} is already eliminated (0 both \textit{Before} and \textit{After}) across datasets, which aligns with GNN4Plan constructing plans over a dependency graph structure that enforces type compatibility.
Despite this, our method continues to reduce non-structural errors such as \textit{Missing Tool} and \textit{Edge Fail}, and it also decreases \textit{Wrong Tool} on several datasets.
Similar to ReAct, DailyLife shows no change in \textit{Wrong Tool} and only a small error mass overall, implying limited headroom for further reductions in that type.

\begin{table}[!h]
\caption{Ablation studies on Dailylife and Ultratool: Node-F1, Link-F1, and Accuracy are reported in \%. The best results are highlighted in boldface, and the second-best results are underlined.}
\label{tab:ablation_results_p2}
\centering
\resizebox{\columnwidth}{!}{
\setlength{\tabcolsep}{4pt}
\renewcommand{\arraystretch}{1.2}
\begin{tabular}{c l ccc ccc}
\toprule
& \multirow{2}{*}{\textbf{Variant}}
& \multicolumn{3}{c}{\textbf{DailyLife}}
& \multicolumn{3}{c}{\textbf{UltraTool}} \\
\cmidrule(lr){3-5} \cmidrule(lr){6-8}
& & \textit{n-F1}$\uparrow$ & \textit{l-F1}$\uparrow$ & \textit{Acc}$\uparrow$
  & \textit{n-F1}$\uparrow$ & \textit{l-F1}$\uparrow$ & \textit{Acc}$\uparrow$ \\
\midrule

\multirow{10}{*}{\rotatebox{90}{\textbf{Direct}}}
& \hspace*{2pt}Raw & \hspace*{2pt}\underline{97.12} & \hspace*{2pt}84.21 & \hspace*{2pt}72.80 & \hspace*{2pt}73.07 & \hspace*{2pt}46.36 & \hspace*{2pt}36.80 \\
& \cellcolor{gray!15} w/o GNN & \cellcolor{gray!15} 95.47 & \cellcolor{gray!15} 84.64 & \cellcolor{gray!15} 74.00 & \cellcolor{gray!15} 72.62 & \cellcolor{gray!15} 45.71 & \cellcolor{gray!15} 34.00 \\
& \hspace*{2pt}w/o Stage-II & \hspace*{2pt}\underline{97.12} & \hspace*{2pt}\underline{87.28} & \hspace*{2pt}\underline{78.60} & \hspace*{2pt}71.33 & \hspace*{2pt}47.40 & \hspace*{2pt}35.60 \\
& \cellcolor{gray!15} w/o Stage-I & \cellcolor{gray!15} \underline{97.12} & \cellcolor{gray!15} \underline{87.28} & \cellcolor{gray!15} \underline{78.60} & \cellcolor{gray!15} 73.33 & \cellcolor{gray!15} 47.56 & \cellcolor{gray!15} 37.60 \\
& \hspace*{2pt}w/o Node Feat. & \hspace*{2pt}\underline{97.12} & \hspace*{2pt}\underline{87.28} & \hspace*{2pt}\underline{78.60} & \hspace*{2pt}\underline{75.15} & \hspace*{2pt}50.49 & \hspace*{2pt}41.00 \\
& \cellcolor{gray!15} w/o Edge Feat. & \cellcolor{gray!15} 96.67 & \cellcolor{gray!15} 86.68 & \cellcolor{gray!15} 76.20 & \cellcolor{gray!15} 75.04 & \cellcolor{gray!15} 50.09 & \cellcolor{gray!15} 41.40 \\
& \hspace*{2pt}w/o Graph FB & \hspace*{2pt}\underline{97.12} & \hspace*{2pt}\underline{87.28} & \hspace*{2pt}\underline{78.60} & \hspace*{2pt}73.97 & \hspace*{2pt}\underline{51.29} & \hspace*{2pt}40.80 \\
& \cellcolor{gray!15} w/o Node FB & \cellcolor{gray!15} \underline{97.12} & \cellcolor{gray!15} \underline{87.28} & \cellcolor{gray!15} \underline{78.60} & \cellcolor{gray!15} 73.20 & \cellcolor{gray!15} \underline{51.29} & \cellcolor{gray!15} \textbf{42.80} \\
& \hspace*{2pt}w/o Edge FB & \hspace*{2pt}\underline{97.12} & \hspace*{2pt}85.13 & \hspace*{2pt}73.80 & \hspace*{2pt}75.13 & \hspace*{2pt}47.06 & \hspace*{2pt}36.80 \\
& \cellcolor{gray!15} Full & \cellcolor{gray!15} \textbf{97.51} & \cellcolor{gray!15} \textbf{87.45} & \cellcolor{gray!15} \textbf{78.76} & \cellcolor{gray!15} \textbf{76.89} & \cellcolor{gray!15} \textbf{52.82} & \cellcolor{gray!15} \textbf{42.80} \\
\midrule

\multirow{10}{*}{\rotatebox{90}{\textbf{ReAct}}}
& \hspace*{2pt}Raw & \hspace*{2pt}\textbf{96.59} & \hspace*{2pt}57.01 & \hspace*{2pt}45.20 & \hspace*{2pt}73.89 & \hspace*{2pt}39.35 & \hspace*{2pt}32.40 \\
& \cellcolor{gray!15} w/o GNN & \cellcolor{gray!15} 94.22 & \cellcolor{gray!15} 62.58 & \cellcolor{gray!15} 51.00 & \cellcolor{gray!15} 74.05 & \cellcolor{gray!15} 39.12 & \cellcolor{gray!15} 32.40 \\
& \hspace*{2pt}w/o Stage-II & \hspace*{2pt}\textbf{96.59} & \hspace*{2pt}83.38 & \hspace*{2pt}68.00 & \hspace*{2pt}74.73 & \hspace*{2pt}50.58 & \hspace*{2pt}42.40 \\
& \cellcolor{gray!15} w/o Stage-I & \cellcolor{gray!15} 96.58 & \cellcolor{gray!15} \underline{85.80} & \cellcolor{gray!15} \underline{76.20} & \cellcolor{gray!15} 73.97 & \cellcolor{gray!15}46.26 & \cellcolor{gray!15}38.40 \\
& \hspace*{2pt}w/o Node Feat. & \hspace*{2pt}\textbf{96.59} & \hspace*{2pt}80.33 & \hspace*{2pt}70.40 & \hspace*{2pt}\underline{76.51} & \hspace*{2pt}\underline{51.90} & \hspace*{2pt}42.20 \\
& \cellcolor{gray!15} w/o Edge Feat. & \cellcolor{gray!15} 94.72 & \cellcolor{gray!15} 83.31 & \cellcolor{gray!15} 70.40 & \cellcolor{gray!15} 76.32 & \cellcolor{gray!15} 51.88 & \cellcolor{gray!15}\textbf{44.20} \\
& \hspace*{2pt}w/o Graph FB & \hspace*{2pt}\textbf{96.59} & \hspace*{2pt}80.83 & \hspace*{2pt}69.60 & \hspace*{2pt}75.24 & \hspace*{2pt}45.28 & \hspace*{2pt}37.40 \\
& \cellcolor{gray!15} w/o Node FB & \cellcolor{gray!15} \textbf{96.59} & \cellcolor{gray!15} 80.56 & \cellcolor{gray!15} 70.40 & \cellcolor{gray!15} 75.73 & \cellcolor{gray!15} 45.75 & \cellcolor{gray!15} 38.00 \\
& \hspace*{2pt}w/o Edge FB & \hspace*{2pt}\textbf{96.59} & \hspace*{2pt}81.40 & \hspace*{2pt}68.60 & \hspace*{2pt}73.89 & \hspace*{2pt}40.05 & \hspace*{2pt}32.40 \\
& \cellcolor{gray!15} Full & \cellcolor{gray!15} \textbf{96.59} & \cellcolor{gray!15} \textbf{85.83} & \cellcolor{gray!15} \textbf{76.35} & \cellcolor{gray!15} \textbf{77.91} & \cellcolor{gray!15} \textbf{52.06} & \cellcolor{gray!15} \textbf{44.20} \\
\midrule

\multirow{10}{*}{\rotatebox{90}{\textbf{GNN4Plan}}}
& \hspace*{2pt}Raw & \hspace*{2pt}\textbf{97.25} & \hspace*{2pt}\underline{87.51} & \hspace*{2pt}\underline{78.80} & \hspace*{2pt}71.68 & \hspace*{2pt}46.99 & \hspace*{2pt}37.80 \\
& \cellcolor{gray!15} w/o GNN & \cellcolor{gray!15} 95.42 & \cellcolor{gray!15} 84.75 & \cellcolor{gray!15} 74.40 & \cellcolor{gray!15} 71.37 & \cellcolor{gray!15} 46.61 & \cellcolor{gray!15} 31.80 \\
& \hspace*{2pt}w/o Stage-II & \hspace*{2pt}\textbf{97.25} & \hspace*{2pt}\textbf{87.61} & \hspace*{2pt}\underline{78.80} & \hspace*{2pt}72.73 & \hspace*{2pt}51.18 & \hspace{2pt}41.20 \\
& \cellcolor{gray!15} w/o Stage-I & \cellcolor{gray!15} \textbf{97.25} & \cellcolor{gray!15} \textbf{87.61} & \cellcolor{gray!15} \underline{78.80} & \cellcolor{gray!15} 71.88 & \cellcolor{gray!15} 47.52 & \cellcolor{gray!15} 38.20 \\
& \hspace*{2pt}w/o Node Feat. & \hspace*{2pt}\textbf{97.25} & \hspace*{2pt}\textbf{87.61} & \hspace*{2pt}\underline{78.80} & \hspace*{2pt}\underline{73.90} & \hspace*{2pt}\underline{52.99} & \hspace*{2pt}43.40 \\
& \cellcolor{gray!15} w/o Edge Feat. & \cellcolor{gray!15} 96.73 & \cellcolor{gray!15} 68.93 & \cellcolor{gray!15} 76.40 & \cellcolor{gray!15} 73.76 & \cellcolor{gray!15} 52.26 & \cellcolor{gray!15}\textbf{43.80} \\
& \hspace*{2pt}w/o Graph FB & \hspace*{2pt}\textbf{97.25} & \hspace*{2pt}\textbf{87.61} & \hspace*{2pt}\underline{78.80} & \hspace*{2pt}72.83 & \hspace*{2pt}52.49 & \hspace*{2pt}42.00 \\
& \cellcolor{gray!15} w/o Node FB & \cellcolor{gray!15} \textbf{97.25} & \cellcolor{gray!15} \textbf{87.61} & \cellcolor{gray!15} \underline{78.80} & \cellcolor{gray!15} 73.55 & \cellcolor{gray!15} 52.56 & \cellcolor{gray!15} 42.20 \\
& \hspace*{2pt}w/o Edge FB & \hspace*{2pt}\textbf{97.25} & \hspace*{2pt}\underline{87.51} & \hspace*{2pt}\underline{78.80} & \hspace*{2pt}72.94 & \hspace*{2pt}47.69 & \hspace*{2pt}37.80 \\
& \cellcolor{gray!15} Full & \cellcolor{gray!15} \textbf{97.25} & \cellcolor{gray!15} \textbf{87.61} & \cellcolor{gray!15} \textbf{78.98} & \cellcolor{gray!15} \textbf{76.44} & \cellcolor{gray!15} \textbf{53.46} & \cellcolor{gray!15} \textbf{43.80} \\
\bottomrule
\end{tabular}
}
\end{table}

\FloatBarrier

\begin{figure*}[t]
  \centering
  \setlength{\tabcolsep}{6pt}
  \begin{tabular}{@{}c@{\hspace{8pt}}c@{\hspace{8pt}}c@{\hspace{8pt}}c@{}}
    \includegraphics[width=0.24\linewidth]{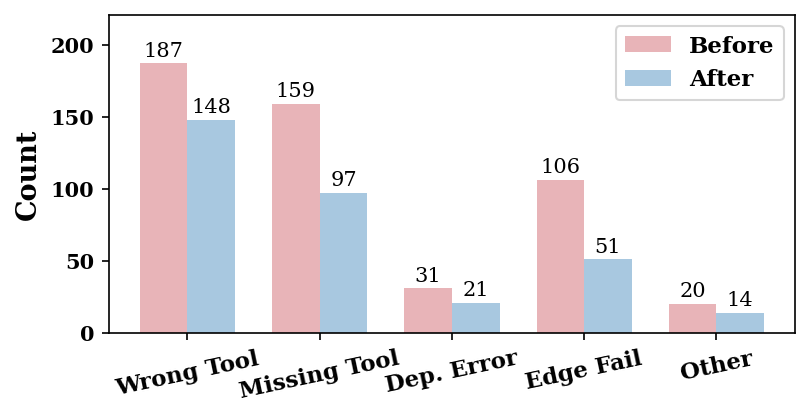} &
    \includegraphics[width=0.24\linewidth]{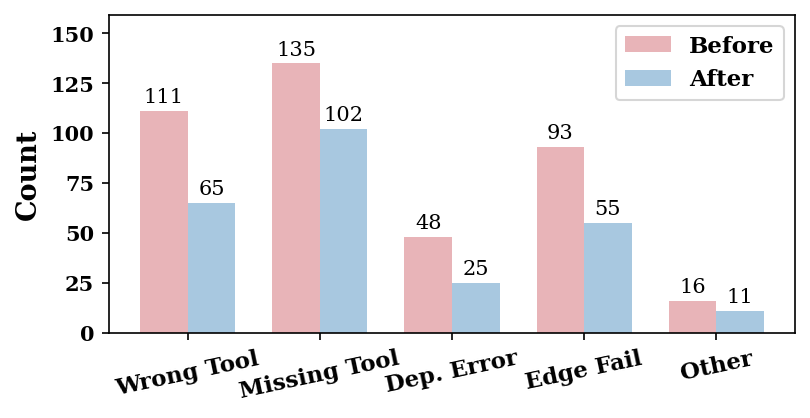} &
    \includegraphics[width=0.24\linewidth]{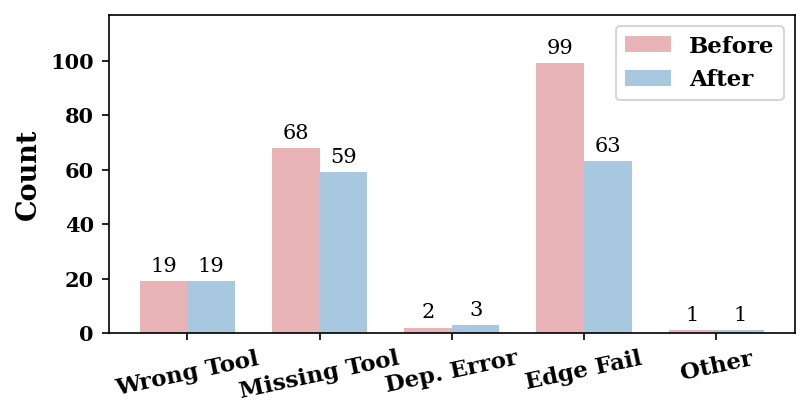} &
    \includegraphics[width=0.24\linewidth]{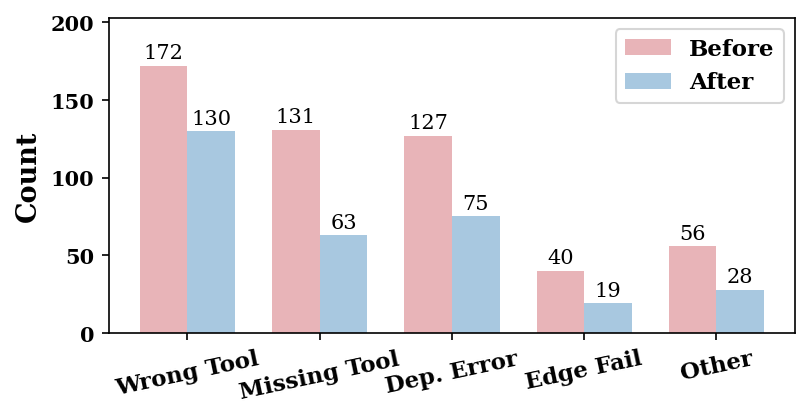}
    \\[0.4em]
    \small \textbf{(a)} \textbf{HuggingFace} & \small \textbf{(b)} \textbf{Multimedia} & \small \textbf{(c)} \textbf{DailyLife} & \small \textbf{(d)} \textbf{UltraTool}
  \end{tabular}
  \caption{Analysis of planning error types before and after correction under ReAct cross datasets with GPT-4o.}
  \label{fig:case_react}
\end{figure*}

\begin{figure*}[t]
  \centering
  \setlength{\tabcolsep}{6pt}
  \begin{tabular}{@{}c@{\hspace{8pt}}c@{\hspace{8pt}}c@{\hspace{8pt}}c@{}}
    \includegraphics[width=0.24\linewidth]{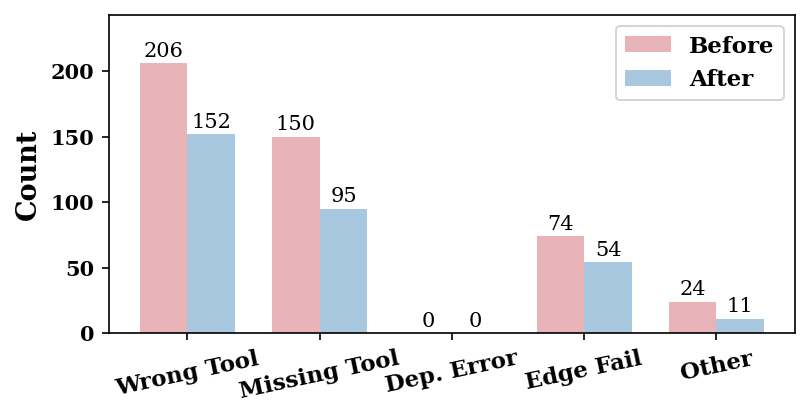} &
    \includegraphics[width=0.24\linewidth]{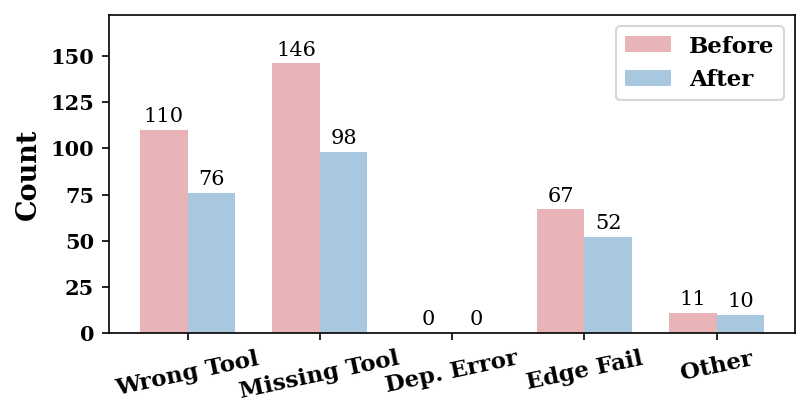} &
    \includegraphics[width=0.24\linewidth]{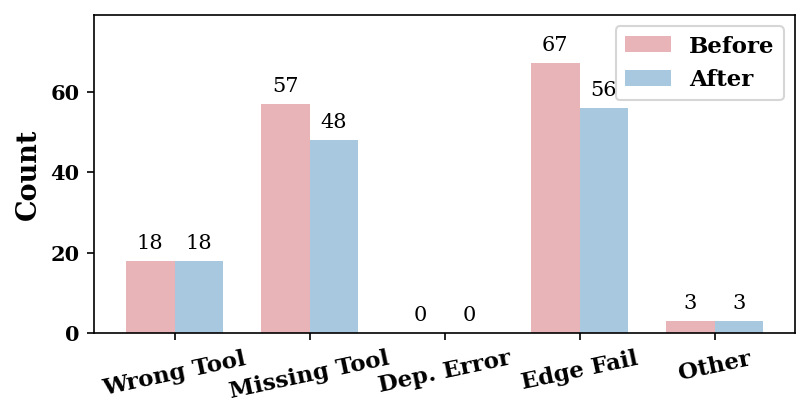} &
    \includegraphics[width=0.24\linewidth]{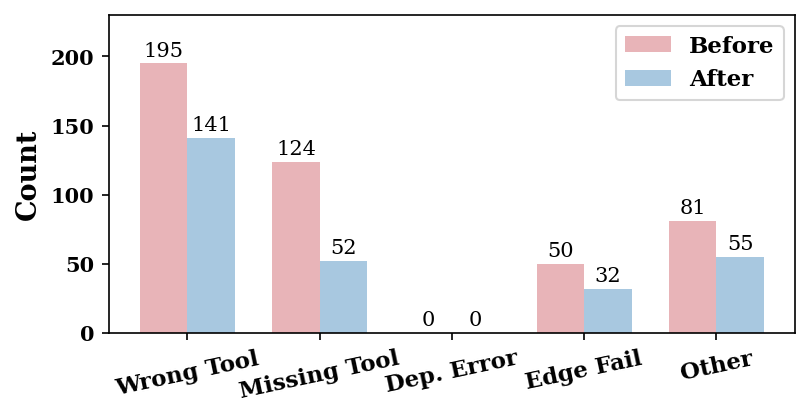}
    \\[0.4em]
    \small \textbf{(a)} \textbf{HuggingFace} & \small \textbf{(b)} \textbf{Multimedia} & \small \textbf{(c)} \textbf{DailyLife} & \small \textbf{(d)} \textbf{UltraTool}
  \end{tabular}
  \caption{Analysis of planning error types before and after correction under GNN4Plan cross datasets with GPT-4o.}
  \label{fig:case_gnn4plan}
\end{figure*}

\newpage
\:
\newpage

\subsection{Additional Visualization of GNN Embeddings}
\label{app:tsne_more}
\begin{figure*}[t]
  \centering
  \setlength{\tabcolsep}{10pt}
  \begin{subfigure}[t]{0.29\linewidth}
    \centering
    \includegraphics[width=\linewidth]{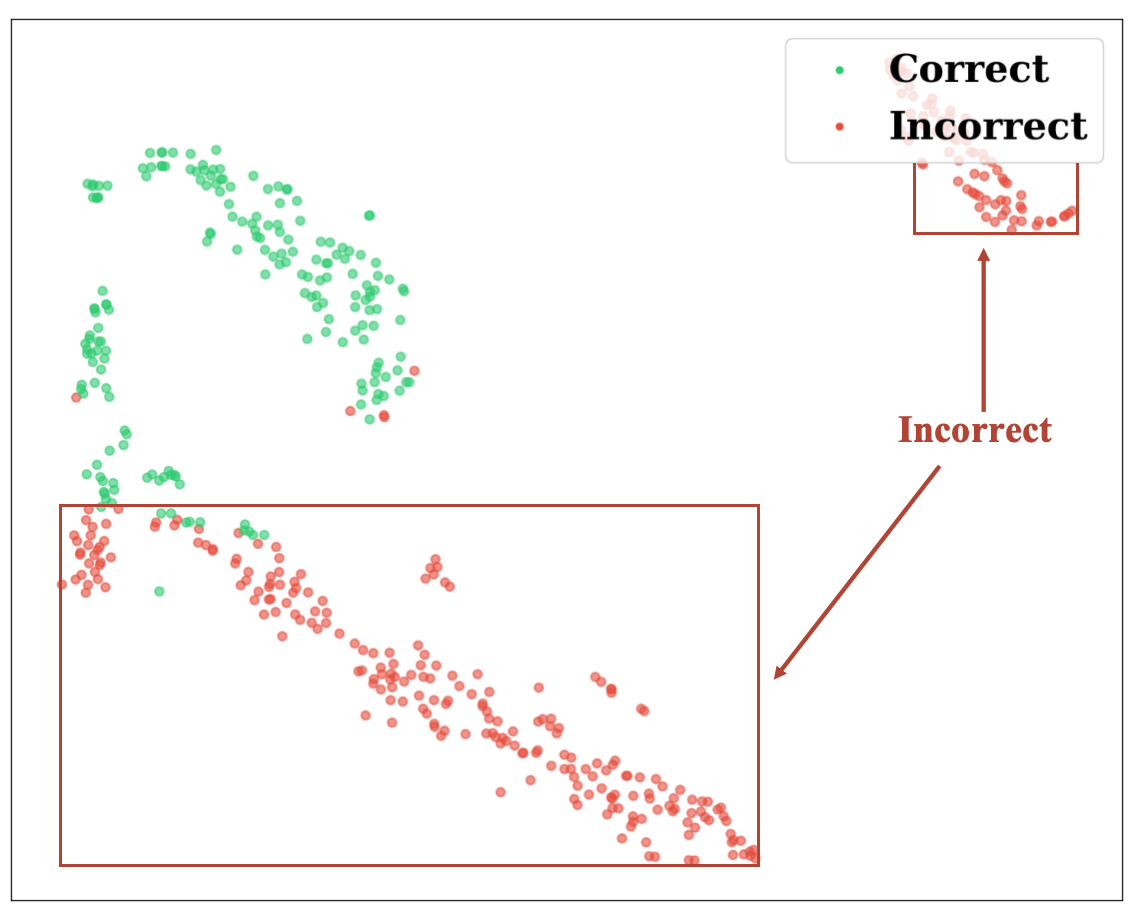}
    \caption{Graph}
    \label{subfig:ultratool_graph_react}
  \end{subfigure}
  \hspace{0.04\linewidth} 
  \begin{subfigure}[t]{0.29\linewidth}
    \centering
    \includegraphics[width=\linewidth]{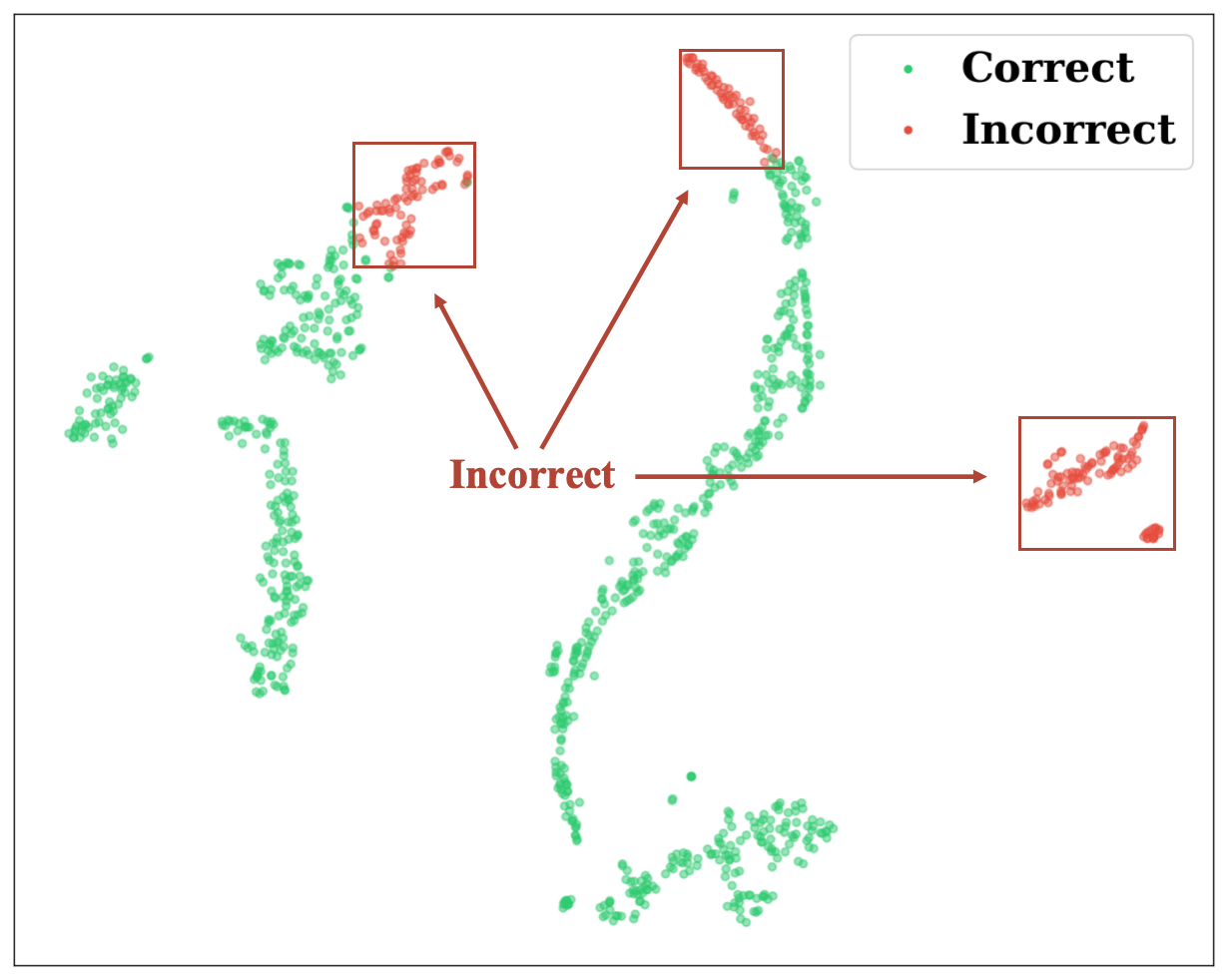}
    \caption{Node}
    \label{subfig:ultratool_node_react}
  \end{subfigure}
  \hspace{0.04\linewidth}
  \begin{subfigure}[t]{0.29\linewidth}
    \centering
    \includegraphics[width=\linewidth]{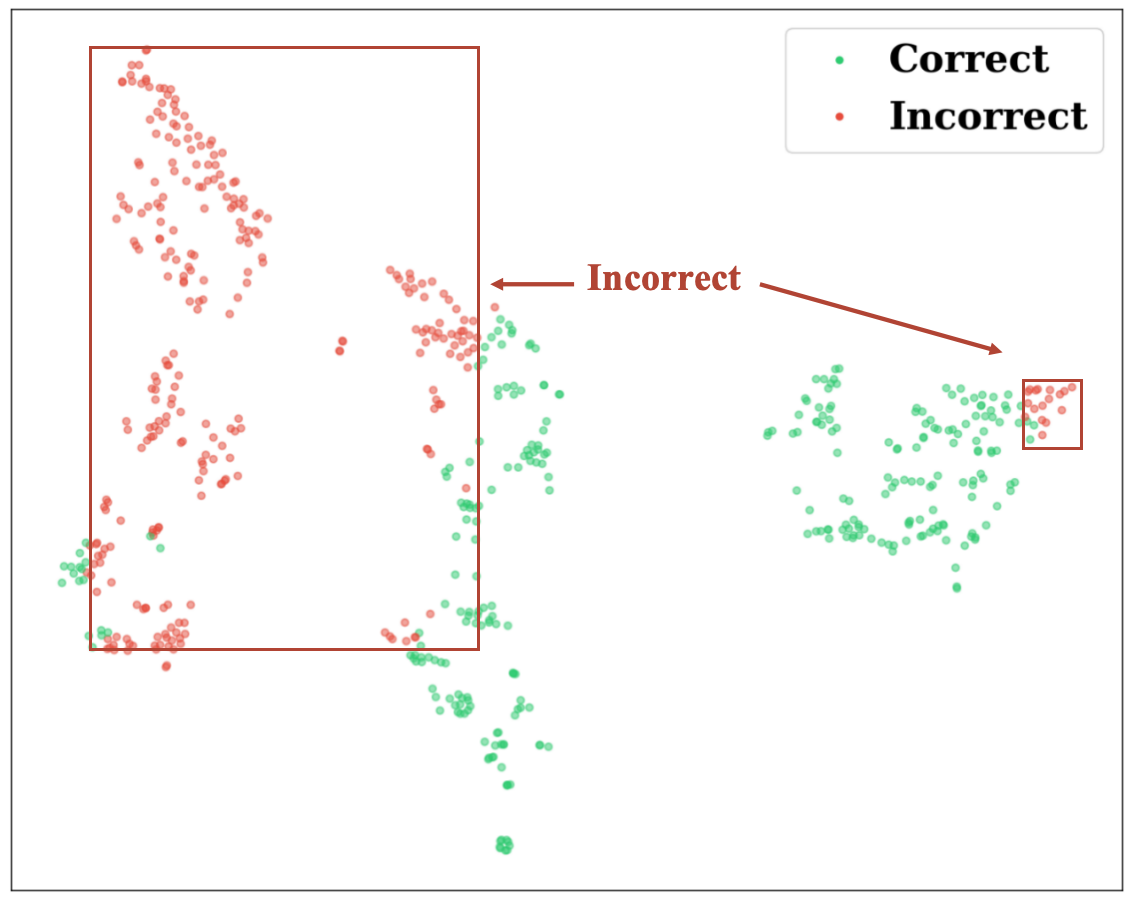}
    \caption{Edge}
    \label{subfig:ultratool_edge_react}
  \end{subfigure}

  \caption{t-SNE visualization of graph-, node-, and edge-level embeddings on UltraTool under ReAct with GPT-4o.}
  \label{fig:ultratool_react}
\end{figure*}

\begin{figure*}[t]
  \centering
  \setlength{\tabcolsep}{10pt} 
  \begin{subfigure}[t]{0.29\linewidth}
    \includegraphics[width=\linewidth]{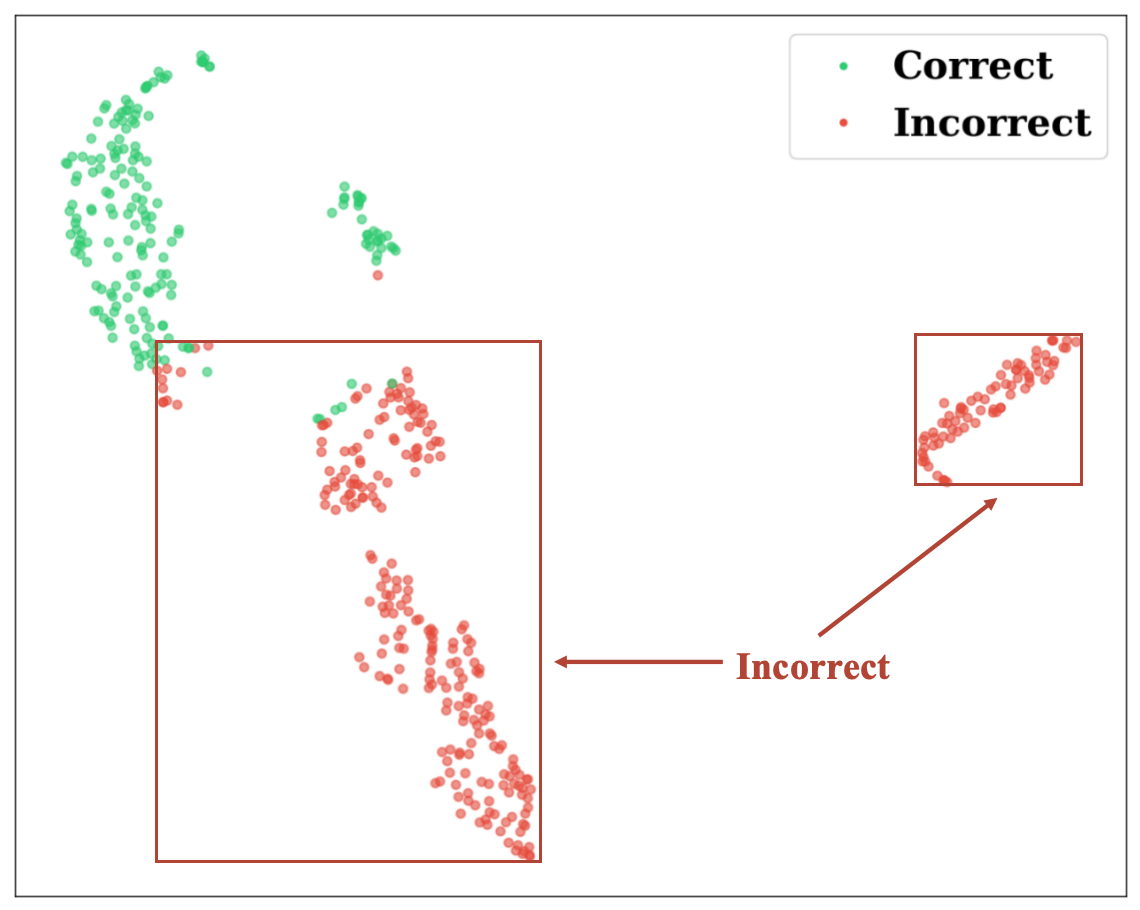}
    \caption{Graph}
    \label{subfig:ultratool_graph_gnn4plan}
  \end{subfigure}
  \hspace{0.04\linewidth}
  \begin{subfigure}[t]{0.29\linewidth}
    \centering
    \includegraphics[width=\linewidth]{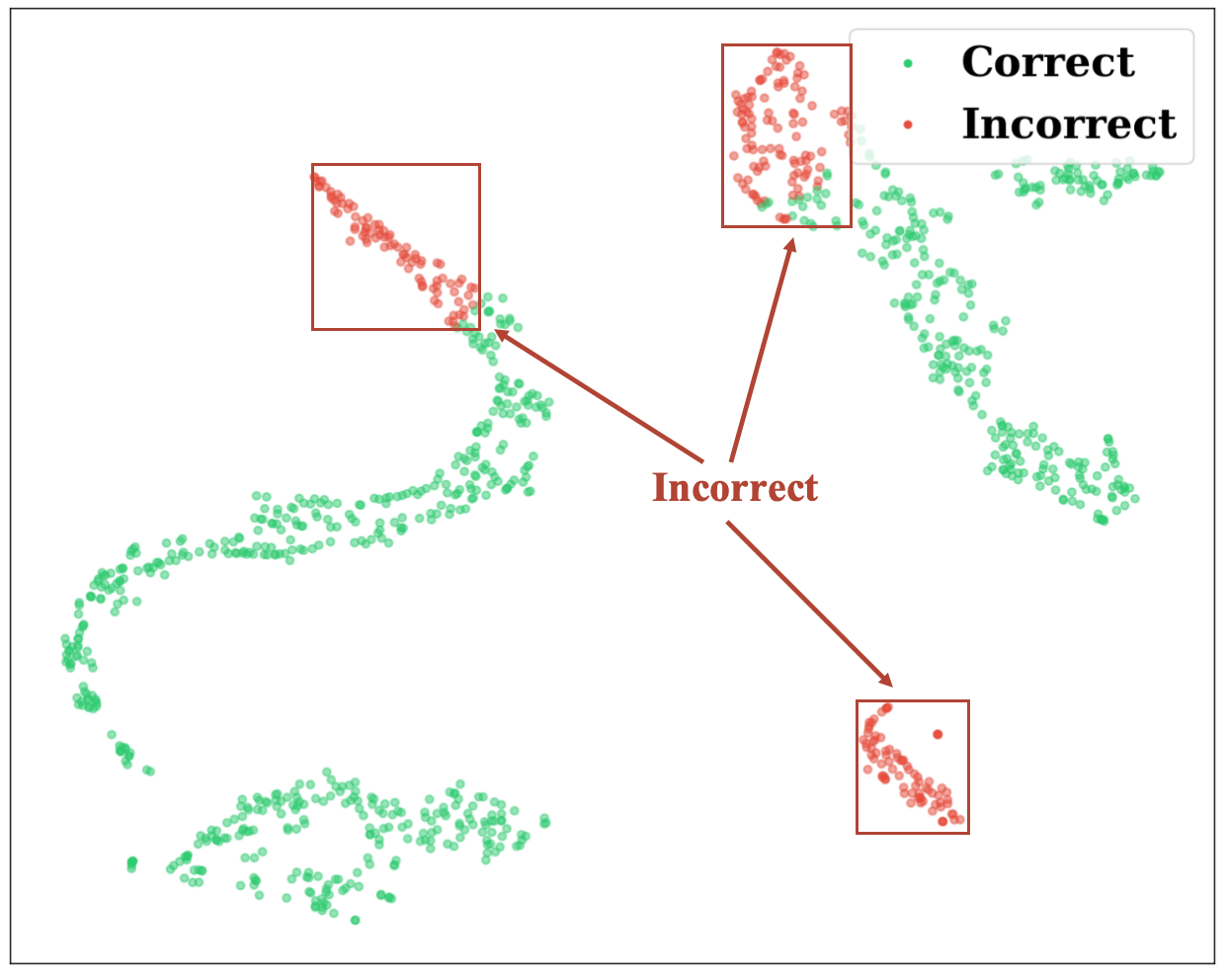}
    \caption{Node}
    \label{subfig:ultratool_node_gnn4plan}
  \end{subfigure}
  \hspace{0.04\linewidth}
  \begin{subfigure}[t]{0.29\linewidth}
    \centering
    \includegraphics[width=\linewidth]{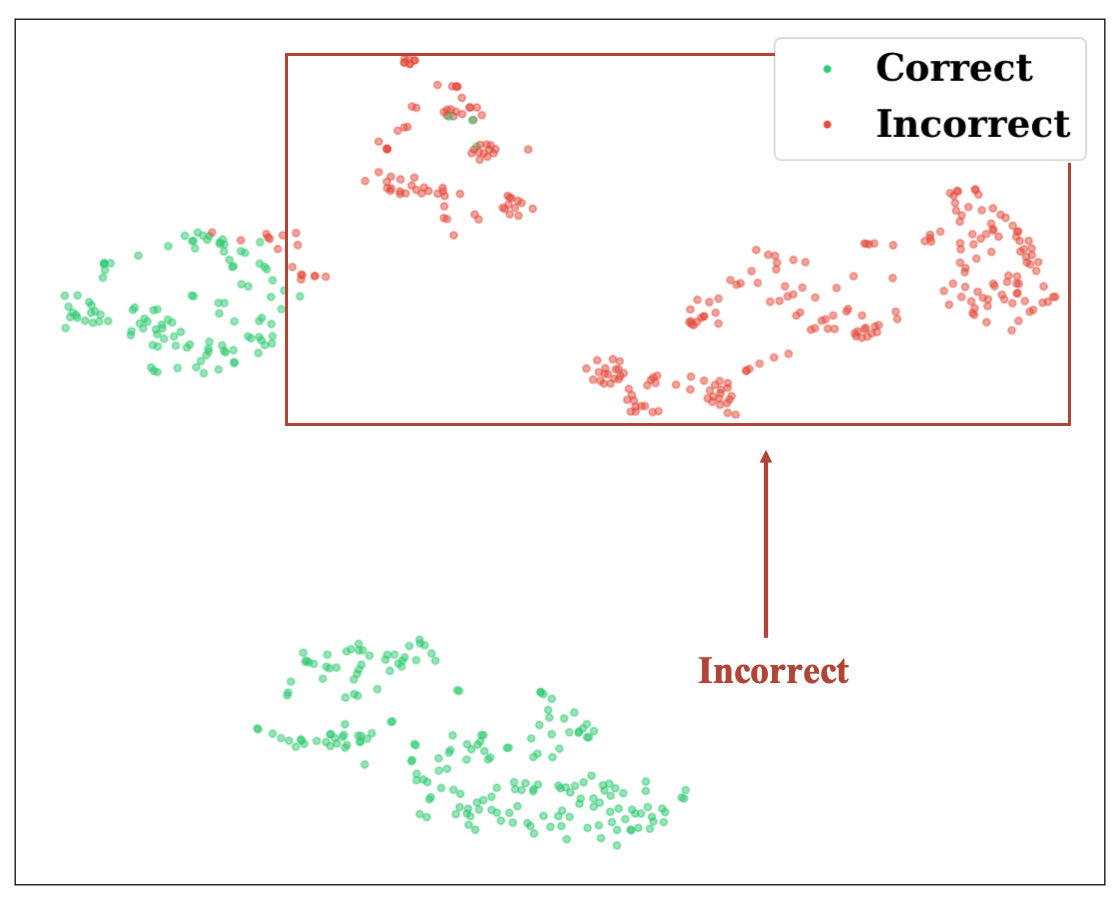}
    \caption{Edge}
    \label{subfig:ultratool_edge_gnn4plan}
  \end{subfigure}

  \caption{t-SNE visualization of graph-, node-, and edge-level embeddings on UltraTool under GNN4Plan with GPT-4o.}
  \label{fig:ultratool_gnn4plan}
\end{figure*}

Figures~\ref{fig:ultratool_react} and~\ref{fig:ultratool_gnn4plan} show the learned embeddings on the largest datase UltraTool under ReAct and GNN4Plan, respectively. 
Similar to the Direct planner (Figure~\ref{fig:ultratool_graph_node_edge}), correct and incorrect samples are largely separable at the graph, node, and edge levels, with limited overlap. This indicates that our verifier learns discriminative embeddings that can be reliably mapped to graph-/node-/edge-level scores, enabling effective correctness discrimination and providing informative signals for acceptance versus correction and for localizing potential errors.

\begin{figure*}[t]
  \centering
  \includegraphics[width=0.32\linewidth]{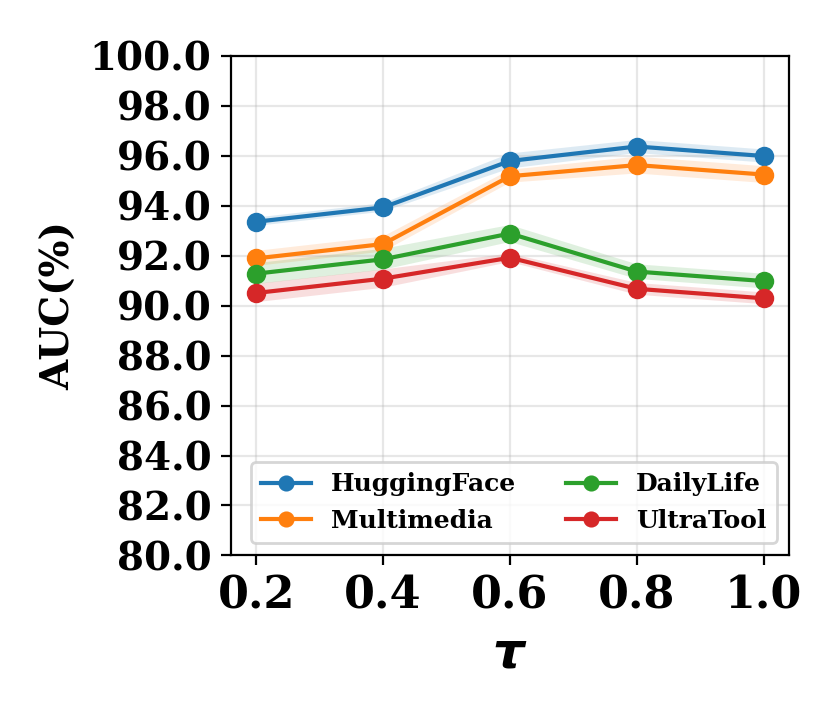}
  \hfill
  \includegraphics[width=0.32\linewidth]{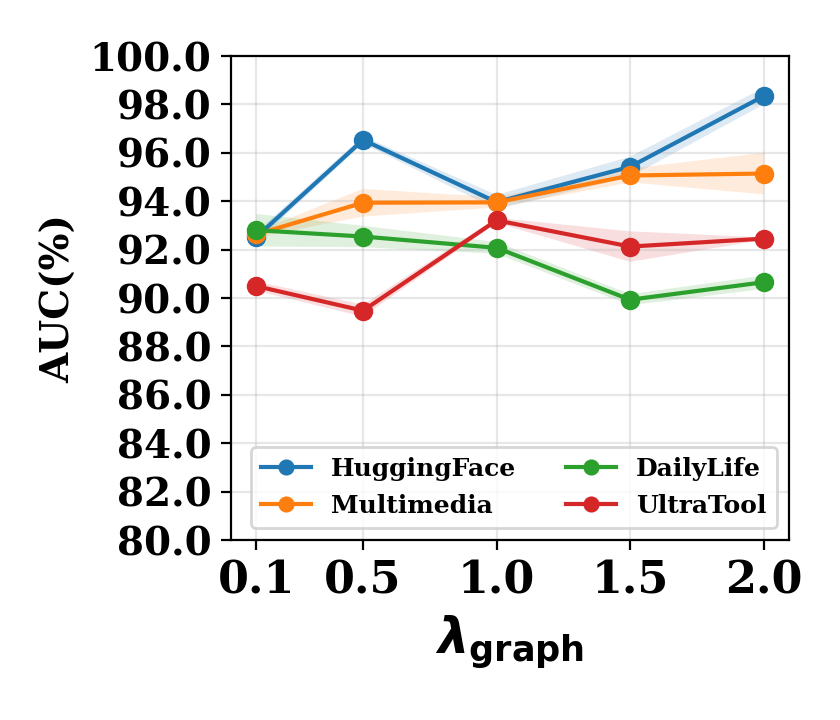}
  \hfill
  \includegraphics[width=0.32\linewidth]{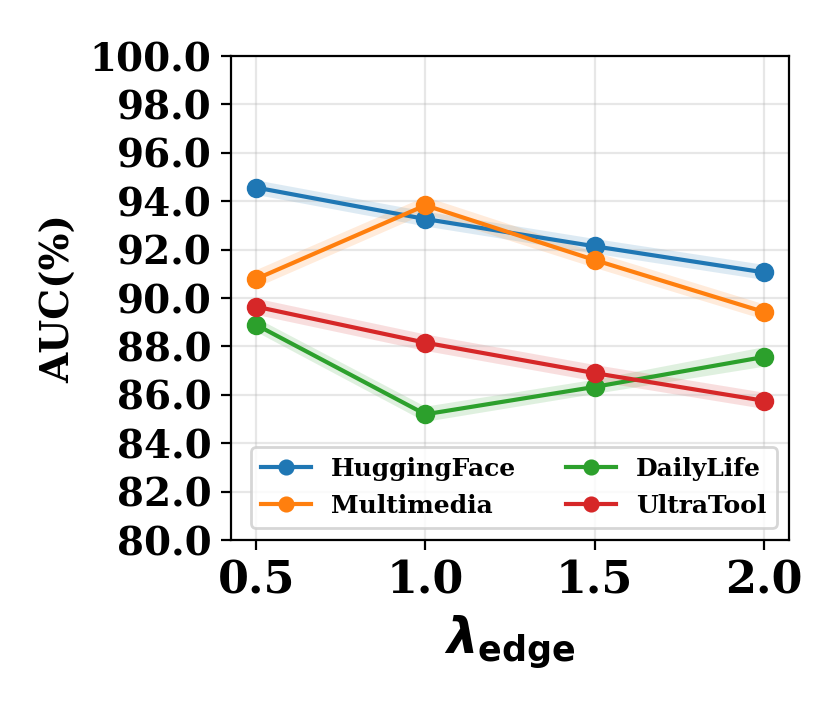}
  \caption{Hyperparameter sensitivity on validation AUC cross datasets with GPT-4o.}
  \Description{Three hyperparameter sensitivity plots showing the effect of tau, lambda graph, and lambda edge on validation AUC across all datasets.}
  \label{fig:hyperparam_sensitivity}
\end{figure*}

\subsection{Hyperparameter Analysis}
\label{sec:hyperparam_analysis}
We tune three key hyperparameters via grid search: the soft target temperature $\tau$, the weight $\lambda_{\text{graph}}$ for the graph objective in Stage~I, and the weight $\lambda_{\text{edge}}$ for the edge objective in Stage~II.
Specifically, we search $\tau\in\{0.2,0.4,0.6,0.8,1.0\}$, $\lambda_{\text{graph}}\in\{0.1,0.5,1.0,1.5,2.0\}$, and $\lambda_{\text{edge}}\in\{0.5,1.0,1.5,2.0\}$.
All hyperparameter analyses are conducted with GPT-4o on the validation split of all datasets.

We select $\tau$ and $\lambda_{\text{graph}}$ by maximizing the validation graph AUC (ROC-AUC) of the graph score, since both mainly affect graph quality learning in Stage~I.
We select $\lambda_{\text{edge}}$ by maximizing a validation local AUC, computed as the average of the ROC-AUC for node risk prediction and edge risk prediction, since it primarily controls the strength of local diagnosis learning in Stage~II.
All AUC values are computed against the self-supervised labels induced by our perturbation operators.

Figure~\ref{fig:hyperparam_sensitivity} shows that $\tau$ typically benefits from moderate values: AUC improves from $\tau=0.2$ to around $0.6$ on all datasets, while larger $\tau$ starts to hurt DailyLife and UltraTool by overly smoothing the distinction between mild and severe perturbations.
For $\lambda_{\text{graph}}$, the trend is more dataset dependent: HuggingFace favors a stronger graph-level regression signal (peaking at larger $\lambda_{\text{graph}}$), whereas DailyLife prefers smaller weights, indicating an interaction between global calibration and the ranking constraint.
In contrast, $\lambda_{\text{edge}}$ exhibits a clearer pattern: increasing $\lambda_{\text{edge}}$ generally degrades AUC on HuggingFace and UltraTool, and Multimedia peaks around $1.0$ but drops sharply for larger values, suggesting that over-weighting the edge objective can dominate training and reduce the discriminability of local diagnosis.
We therefore select the best configuration for each dataset based on its validation performance, and keep it fixed in all main experiments.

\newpage
\null\newpage
\null\newpage

\subsection{Case Study}
\label{app:case_study}
\begin{figure}[t]
  \centering
  \includegraphics[width=\linewidth]{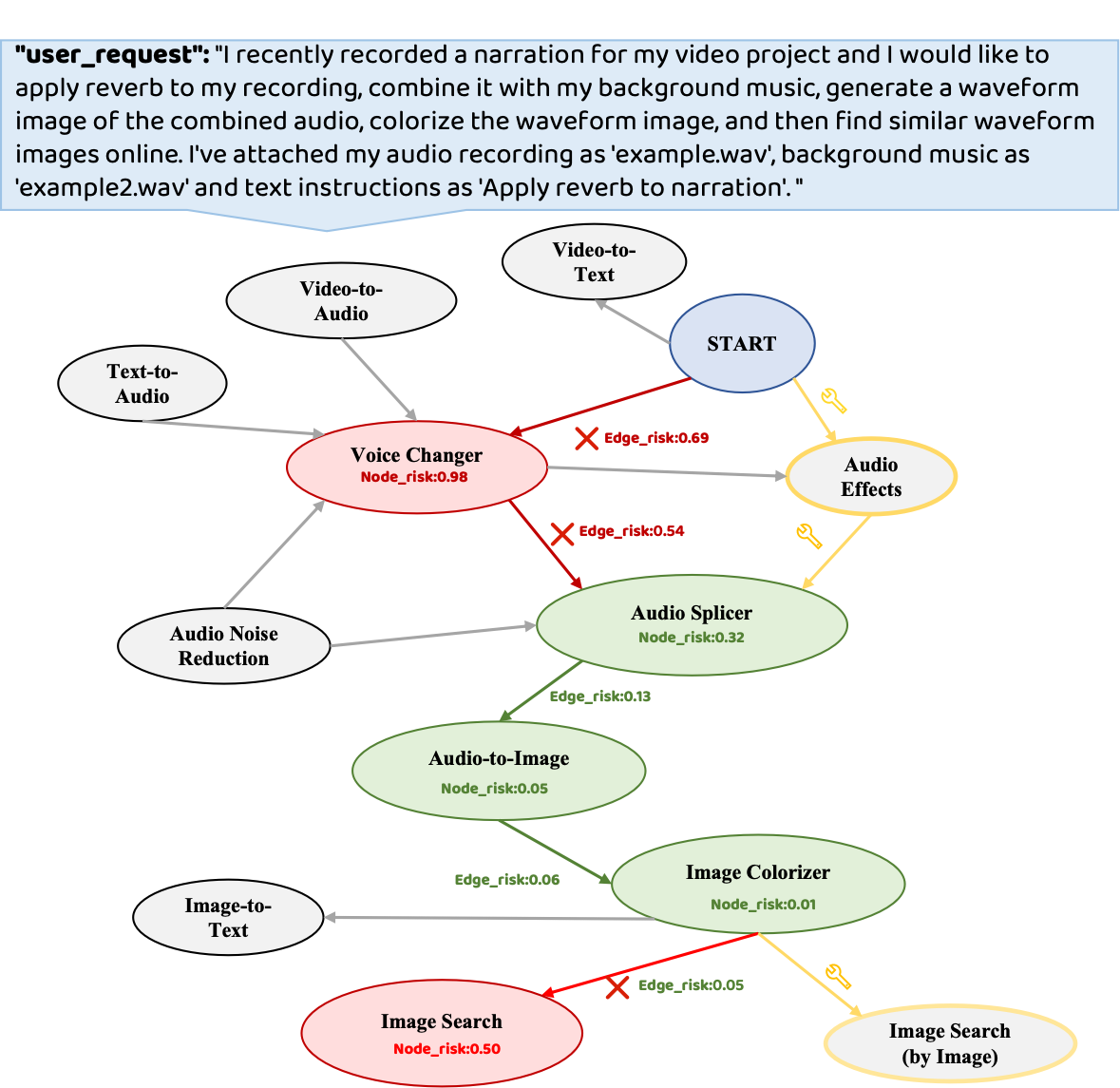}
  \caption{Case study on Multimedia.}
  \label{fig:case_multimedia}
\end{figure}

Figure~\ref{fig:case_multimedia} presents an example from the Multimedia dataset with the Direct planner. The request asks to apply reverb to a narration, mix it with background music, generate a waveform image of the combined audio, colorize the waveform image, and then retrieve similar waveform images online. In the visualization, green nodes denote correctly selected tools by the planner, red nodes denote incorrect selections, gray nodes are candidate tools provided by the dependency graph, and yellow nodes are the corrected tools chosen from the candidates; the node- and edge-level risk scores predicted before correction are also annotated on the corresponding nodes and edges.
Our method corrects two representative failure modes in this case: a confusable tool choice at the first step and an invalid downstream dependency. Guided by the verifier's risk scores, we replace the mistaken tool with a semantically closer alternative from its similar tool neighborhood, and we simultaneously correct the incorrect transition by selecting a type-compatible retrieval tool under the dependency graph constraints, resulting in an executable plan that better matches the user request.

\null\newpage

\end{document}